\pdfoutput=1
\RequirePackage{ifpdf}
\ifpdf 
\documentclass[pdftex]{sigma}
\else
\documentclass{sigma}
\fi

\begin{document}

\allowdisplaybreaks

\renewcommand{\PaperNumber}{075}

\FirstPageHeading

\ShortArticleName{Image Sampling with Quasicrystals}

\ArticleName{Image Sampling with Quasicrystals}

\Author{Mark GRUNDLAND~$^\dag$, Ji\v r\'\i\ PATERA~$^\ddag$, Zuzana MAS\'AKOV\'A~$^\S$ and Neil A. DODGSON~$^\dag$}

\AuthorNameForHeading{M. Grundland, J. Patera, Z. Mas\'akov\'a and N.A. Dodgson}

\Address{$^\dag$~Computer Laboratory, University of Cambridge, UK}
\EmailD{\href{mailto:Mark@Eyemaginary.com}{Mark@Eyemaginary.com}, \href{mailto:Neil.Dodgson@cl.cam.ac.uk}{Neil.Dodgson@cl.cam.ac.uk}}

\Address{$^\ddag$~Centre de Recherches Math\'ematiques, Universit\'e de Montr\'eal, Canada}
\EmailD{\href{mailto:patera@crm.umontreal.ca}{patera@crm.umontreal.ca}}

\Address{$^\S$~Department of Mathematics FNSPE, Czech Technical
University in Prague, Czech Republic}
\EmailD{\href{mailto:zuzana.masakova@fjfi.cvut.cz}{zuzana.masakova@fjfi.cvut.cz}}

\ArticleDates{Received December 15, 2008, in f\/inal form July 06, 2009;  Published online July 20, 2009}

\Abstract{We investigate the use of quasicrystals in image sampling. Quasicrystals produce space-f\/illing, non-periodic point sets that are uniformly discrete and relatively dense, thereby ensuring the sample sites are evenly spread out throughout the sampled image. Their self-similar structure can be attractive for creating sampling patterns endowed with a decorative symmetry. We present a brief general overview of the algebraic theory of cut-and-project quasicrystals based on the geometry of the golden ratio. To assess the practical utility of quasicrystal sampling, we evaluate the visual ef\/fects of a variety of non-adaptive image sampling strategies on photorealistic image reconstruction and non-photorealistic ima\-ge rendering used in multiresolution image representations. For computer visualization of point sets used in image sampling, we introduce a mosaic rendering technique.}


\Classification{20H15; 52C23; 68U99; 82D25}

\section{Introduction}

\begin{figure}[p]\small
\centering
\begin{tabular}{@{}cc@{}}
\includegraphics[width=76mm]{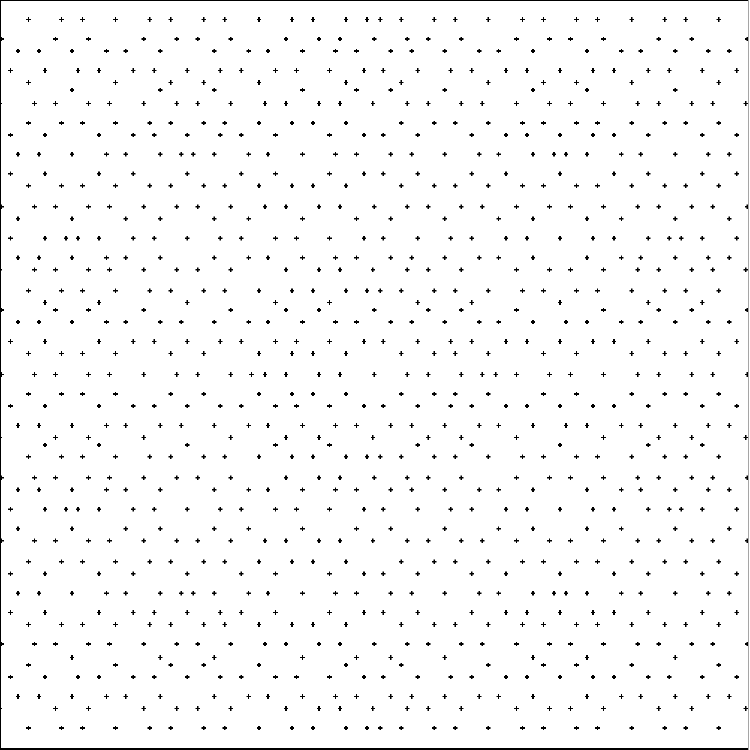}
 &
\includegraphics[width=76mm]{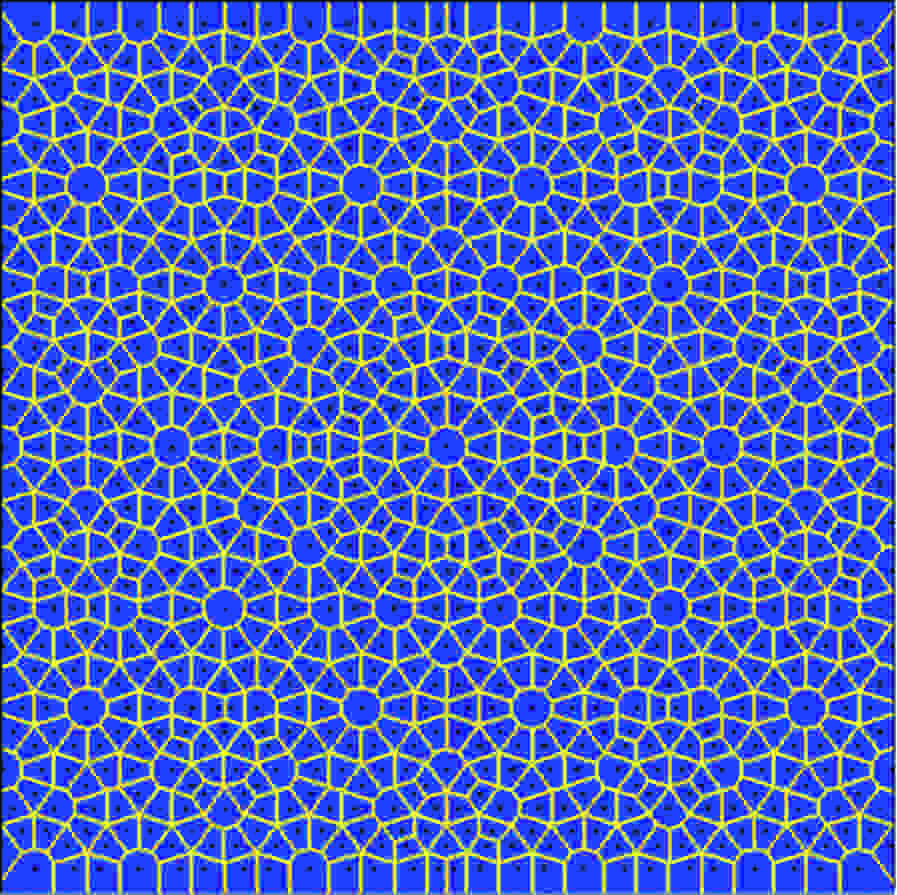}
\\
Point set & Voronoi diagram \\[3mm]
\includegraphics[width=76mm,height=76mm]{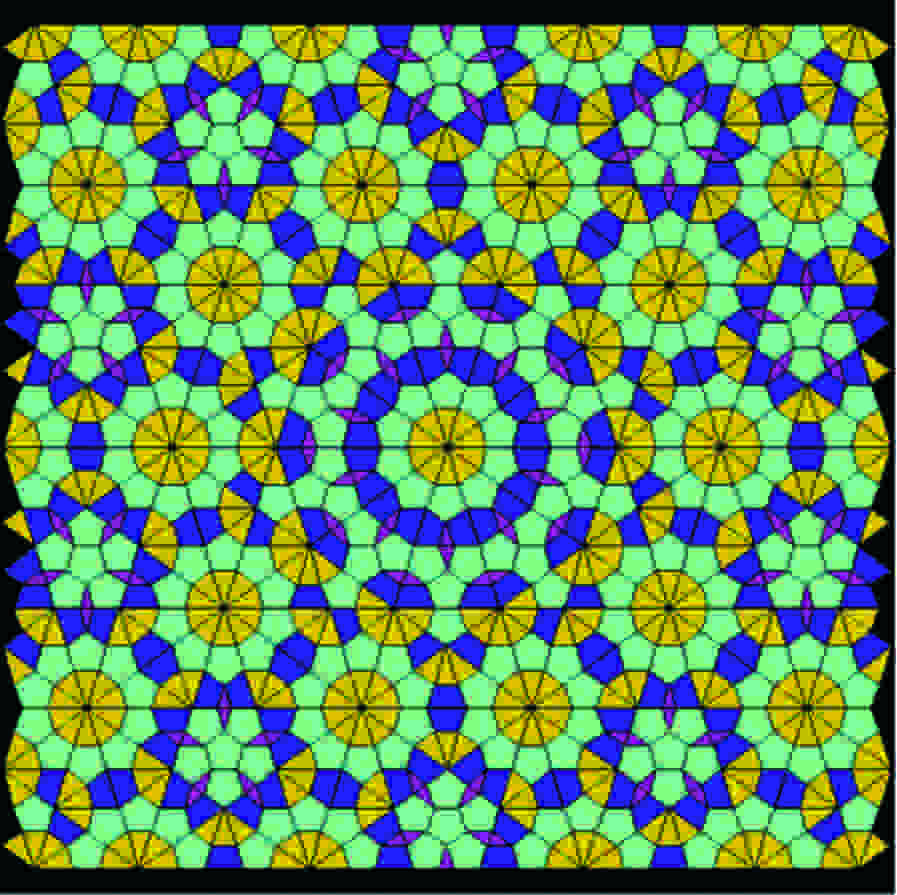}
 &
\includegraphics[width=76mm,height=76mm]{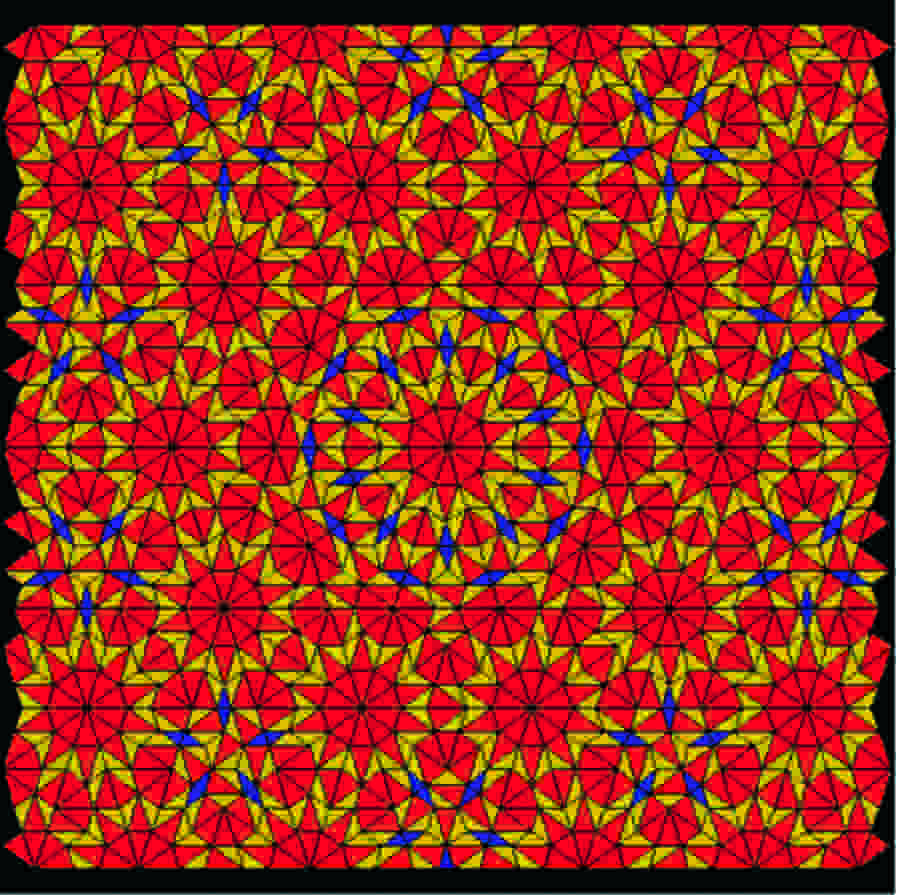}
\\
Delaunay graph & Delaunay triangulation
\end{tabular}
\caption{Quasicrystal tilings produced using spatial proximity
graphs. In these visualizations, a non-periodic, rotationally
symmetric point set (top left) is depicted as a planar tiling
induced by a Voronoi diagram (top right), a Delaunay graph (bottom
left), and a Delaunay triangulation (bottom right). This set of
1035 points comprises a cut-and-project quasicrystal derived from
the standard root lattice of the non-crystallographic Coxeter
group~$H_2$. Its viewing window is a square centered at the origin
with radius~1, while its acceptance window is a decagon centered
at the origin with radius $\tau^5 + \tau^3$, where $\tau$  is the
golden ratio. The visualization of the quasicrystal tilings
reveals some remarkable properties. It is well known that
quasicrystals can exhibit f\/ive and ten fold rotational symmetry,
an impossible feat for any periodic tiling. Recently, it has been
shown analytically that a quasicrystal Delaunay graph can yield a
non-periodic tiling with four distinct tile shapes~\cite{ZichIII}.
As illustrated by this visualization, a Delaunay triangulation of
a cut-and-project quasicrystal can yield a non-periodic tiling
with just three distinct tile shapes. }\label{Fig1}\vspace{-2mm}
\end{figure}

Non-periodic tilings have emerged as an important mathematical
tool in a variety of computer graphics applications
\cite{Tile-based methods for interactive applications}. They have
proven especially useful in the design of sampling algorithms,
where they serve to direct the spatial distribution of rendering
primitives by enforcing spatial uniformity while precluding
regular repetition. Recently, Wang tilings~\cite{WangTiles,Tiled
blue noise samples,Recursive wang tiles for real-time blue noise,A
procedural object distribution function,AlternativeForWang},
Penrose tilings~\cite{Fast hierarchical importance sampling with
blue noise properties}, Socolar tilings
\cite{BuildingLowDiscrepancy} and polyominoes~\cite{Sampling with
polyominoes} have been used to generate point sets for
non-periodic sampling. In one of the earliest applications of
non-periodic tilings in computer graphics, Penrose tilings
\cite{Penrose tiling,Tilings and patterns,Quasicrystals and
geometry} were employed by Rangel-Mondragon and Abas
\cite{Computer generation of penrose tilings} in the design of
decorative patterns inspired by Islamic art. They had ef\/fectively
reinvented the medieval trade secrets of the craftsmen of
f\/ifteenth century Islamic mosques~\cite{MedievalArchitecture} who
crea\-ted by hand highly intricate mosaics closely resembling
quasicrystal tilings only discovered by modern science in the late
twentieth century. Wang tilings~\cite{Penrose tiling,Tilings and
patterns} were f\/irst introduced by Stam~\cite{Aperiodic texture
mapping} in order to enable wave texture patches to cover water
surfaces of arbitrary size without the appearance of regularly
repeating artifacts. Further computer graphics applications of
non-periodic tilings include texture mapping and
synthesis~\cite{WangTiles,Recursive wang tiles for real-time blue
noise,A procedural object distribution
function,AlternativeForWang,Aperiodic texture mapping,Tile-based
texture mapping on graphics hardware}, photorealistic rendering
using environmental maps~\cite{A procedural object distribution
function,Fast hierarchical importance sampling with blue noise
properties}, and non-photorealistic rendering using stippling~\cite{Recursive wang tiles for real-time blue noise,A procedural
object distribution function}. For computer graphics applications,
non-periodic tilings have usually been generated by geometric
constructs, such as matching rules and hierarchical substitution
\cite{Aperiodic tiling,Tilings and patterns}. In this work, we
present the cut-and-project method of generating quasicrystals as
an alternative algebraic approach to the production of
non-periodic tilings and point sets (Fig.~\ref{Fig1}).
This algebraic approach has the advantages of being
straightforward to implement, easy to calculate, and readily
amenable to rigorous mathematical analysis. Moreover, it may be
directly extended to higher dimensions as well as adaptive
sampling applications, although this is outside the scope of our
present work. We choose to base our method on the algebra of the
golden ratio $\tau =\frac12(1+\sqrt5)$, as its geometrical
properties have been previously successfully exploited in computer
graphics in the context of spatial sampling~\cite{Fast
hierarchical importance sampling with blue noise properties} and
color quantization~\cite{Color quantization and processing by
fibonacci lattices} techniques that rely on the Fibonacci number
system. For an introduction to the theory of quasicrystals,
consult Senechal's comprehensive textbook~\cite{Quasicrystals and
geometry}, while a more advanced treatment of the cut-and-project
method may be found in surveys by Patera
\cite{Non-crystallographic root systems and quasicrystals} as well
as Chen, Moody, and Patera~\cite{Non-crystallographic root
systems}.

In the evaluation of the ef\/fectiveness of quasicrystals as a
non-adaptive image sampling stra\-te\-gy, our work is motivated by the
use of image sampling in multiresolution image representation and
progressive image rendering. In particular, we base our
experimental investigations on our experience with the development
of a point-based rendering approach to multiresolution image
representation for digital
photography~\cite{StyleAndContent,Stylized multiresolution image
representation} based on scattered data interpolation techniques~\cite{ScatteredData},
which has been shown to support a secure and
compact image encoding suitable for both photorealistic image
reconstruction and non-photorealistic image rendering. A thorough
discussion of the standard image sampling strategies can be found
in Glassner's textbook~\cite{Principles of digital image
synthesis}. Their ef\/fectiveness has been extensively investigated
for use in photorealistic computer graphics applications
\cite{Discrepancy as a quality measure for sample distributions},
such as Monte Carlo integration in 3D ray tracing. For
non-adaptive sampling, the key trade-of\/f is between aliasing and
noise, as exemplif\/ied by the regular structure of periodic
sampling using a~square grid and the irregular clustering of
random sampling using a~uniform distribution. The classic
compromise strategies are jittered sampling
\cite{StochasticSampling,Antialiasing through stochastic
sampling,Filtered jitter}, which disrupts the regularity of
periodic sampling by randomly perturbing the sample sites, and
quasirandom sampling~\cite{Numerical recipes in c}, which avoids
the irregularity of random sampling by ensuring a consistent
density of sampling is maintained. In our experiments, we
demonstrate that quasicrystal sampling can permit more accurate
photorealistic image reconstruction than either standard jittered
sampling or standard quasirandom sampling. For photorealistic
image reconstruction~\cite{Principles of digital image synthesis},
the ideal strategy is generally considered to be the Poisson disk
distribution~\cite{StochasticSampling,Antialiasing through
stochastic sampling}, random point sets conditioned on a minimum
distance between the points, while for non-photorealistic image
rendering~\cite{FloatingPoints,Simulating decorative
mosaics,Pointillist halftoning,BeyondStippling,Weighted voronoi
stippling}, a popular strategy relies on centroidal Voronoi
diagrams~\cite{CentroidalVoronoiTesselations}, optimized point
sets with every point placed at the centroid of its Voronoi
polygon. As both of these sampling strategies prove time consuming
to compute exactly, a variety of approximation techniques have
been proposed to produce sampling patterns that have similar
properties in the frequency domain, in particular those that
exhibit a blue noise Fourier power spectrum cha\-rac\-te\-ristic of a~Poisson disk distribution.
Historically, blue noise sampling
strategies relied on slow, trial and error, stochastic procedures
involving dart throwing algorithms that approximate the Poisson
disk distribution by rejecting prospective locations for new
sample sites whenever they are deemed to be too close to the
preceding samples~\cite{StochasticSampling,Antialiasing through
stochastic sampling,Hierarchical poisson disk sampling
distributions,Spectrally optimal sampling for distribution ray
tracing}. Improved performance of blue noise sampling can be
obtained through the use of ef\/f\/icient geometric data structures
\cite{SpatialDataStructuren,FarthestPointStrategy,EfficientGeneration,Poisson
disk point sets by hierarchical dart throwing} and parallel
processing GPU hardware~\cite{Parallel poisson disk sampling}.
Alternatively, one can readily generate a blue noise sampling
pattern using a non-periodic tiling composed of a suitable set of
tiles, where each tile contains a precomputed optimal arrangement
of sample sites~\cite{WangTiles,Tiled blue noise samples,Recursive
wang tiles for real-time blue noise,A procedural object
distribution function,AlternativeForWang,Fast hierarchical
importance sampling with blue noise properties,Sampling with
polyominoes}. A~detailed evaluation of the spectral properties of
various blue noise sampling algorithms can be found in the survey
by Lagae and Dutre~\cite{ComparisonOfMethods}. In practical
applications, it can often be quite dif\/f\/icult to visually
distinguish between renditions produced by blue noise sampling
patterns generated using dif\/ferent algorithms. Hence, for the
purpose of our evaluation, we relied on farthest point
sampling~\cite{FarthestPointStrategy} since in previous work we
have shown this blue noise sampling technique to be highly
suitable for multiresolution image representation~\cite{Stylized
multiresolution image representation}. In general, blue noise
sampling algorithms tend to have higher requirements for either
computational processing, data storage, or implementation
complexity than simpler sampling strategies, such as quasicrystal
sampling. Naturally, simpler sampling strategies cannot replicate
all of the desirable qualities of blue noise sampling.
Nevertheless, as demonstrated by our evaluation, quasicrystal
sampling is shown to be prof\/icient at supporting a uniform
sampling density, centroidal Voronoi regions, accurate image
reconstruction, and progressive image rendering, despite having
only a small number of local sampling conf\/igurations arranged in
an anisotropic manner incompatible with a blue noise Fourier power
spectrum. Hence, as a potential alternative to periodic sampling
in image representation, sampling using cut-and-project
quasicrystals can deterministically guarantee minimum and maximum
distances between nearest neighbors in a uniformly space-f\/illing
sampling pattern without the overhead of geometric data structures
or tiling lookup tables for tracking their placement.

\section{Method}

While a periodic point set is characterized by its translational
symmetries, a non-periodic point set admits no translational
symmetries. For use in image sampling, we focus on non-periodic
point sets that are determined by their inf\/lation symmetries. In
such a non-periodic point set, a f\/ixed conf\/iguration of sample
sites can be repeated at dif\/ferent scales to generate a
self-similar pattern. The simplest way of producing non-periodic
point sets is to use hierarchical substitution
tilings~\cite{Aperiodic hierarchical tilings,Tilings and
patterns}. For instance, hierarchical substitution can be readily
applied to the famous Penrose tiling~\cite{Penrose tiling,Tilings
and patterns,Fast hierarchical importance sampling with blue noise
properties,Computer generation of penrose tilings,Quasicrystals
and geometry}. The strategy starts with a small set of polygonal
tiles. The tile set is carefully designed so that each tile can be
decomposed by geometric subdivision into smaller instances of
itself and the other tiles. A hierarchy is formed whereby an
existing tile becomes the parent of new child tiles. Starting with
an initial conf\/iguration of the tiles scaled to cover the image
plane, the tiling is ref\/ined through an iterative process of
def\/lation and substitution. The tile vertices or centroids are
used to derive a point set from the tiling. The choice of the
initial conf\/iguration appears mirrored in the global structure and
symmetry properties of the tiling and the resulting point set.

\begin{figure}[t]
\small
\setlength{\fboxsep}{0pt}
\setlength{\fboxrule}{0.75pt}
\begin{center}
\begin{tabular}{cc}
 \framebox{\includegraphics[width=74mm]{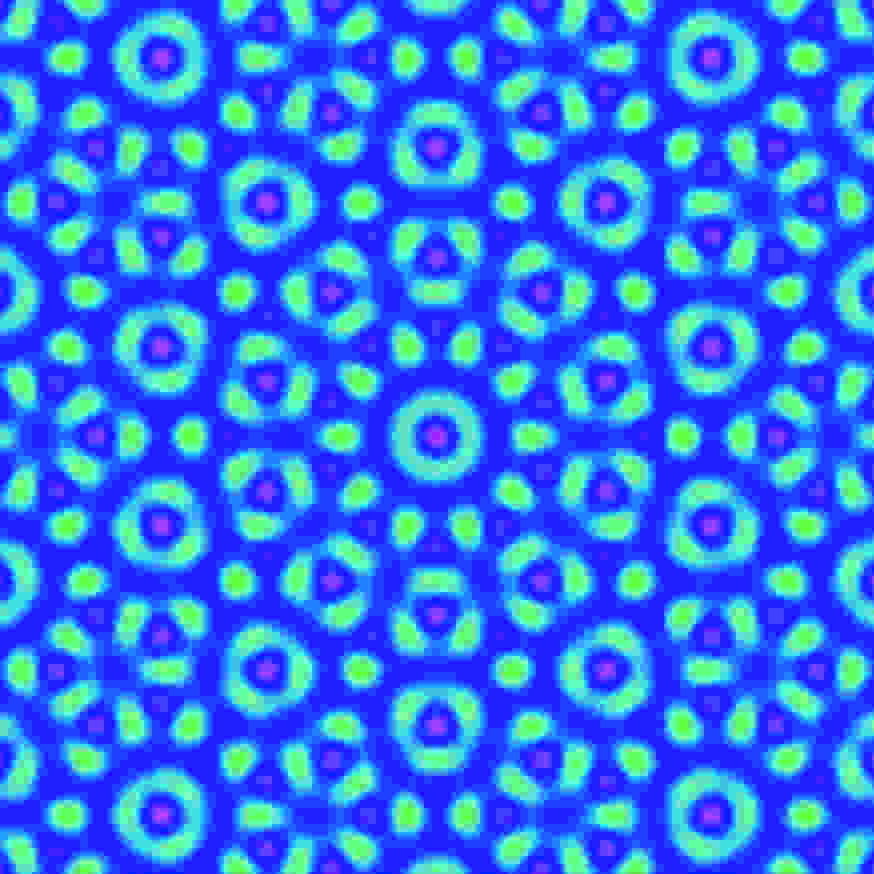}}
&
\framebox{\includegraphics[width=74mm]{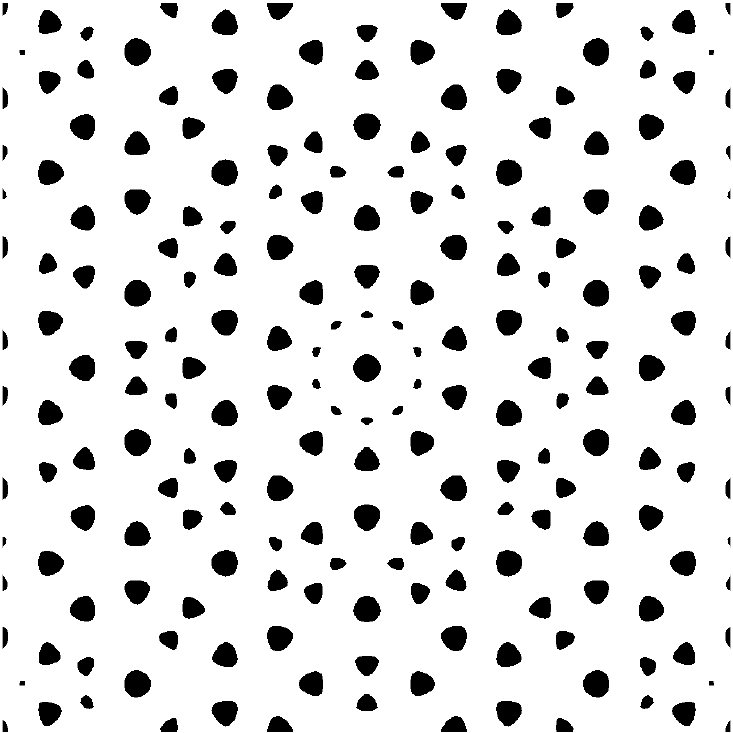}}\\
Phase function contour map & Phase function level set\\[2mm]
\end{tabular}
{\setlength{\fboxsep}{3pt}\setlength{\fboxrule}{0.5pt}
\framebox{
\parbox[c]{0.148\textwidth}{\includegraphics[scale=0.3]{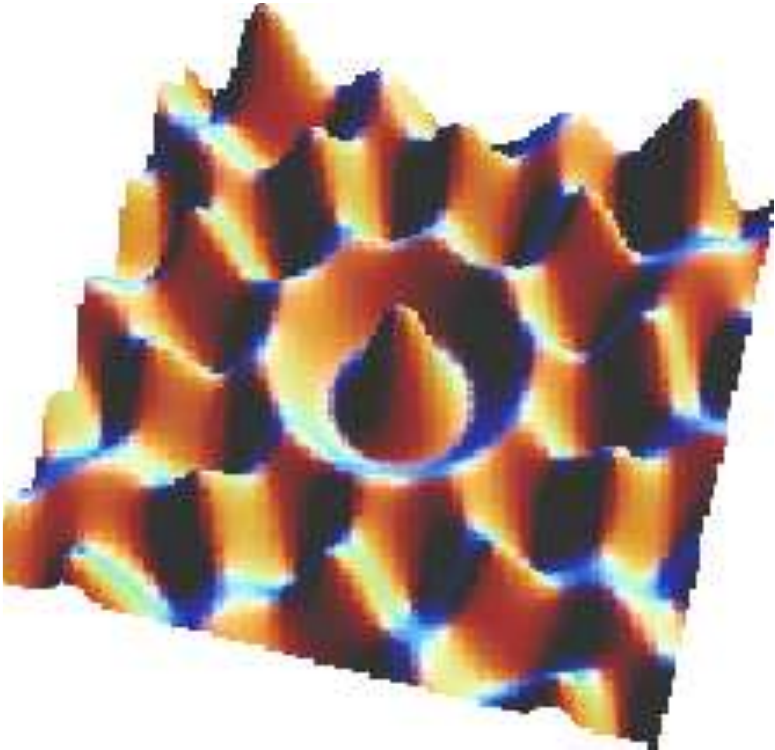}} 
\quad
\parbox[c]{0.78\textwidth}{
{\bf Quasicrystal phase function:} \ $f(z)=\sum _{j=0}^{9}\exp\big(2\pi i(\zeta ^{j} )\cdot (2\tau ^{4} z)\big)$ \
def\/ined \\ for regular decagon vertices $\zeta ^{j}
=\exp\big({\textstyle\frac{2\pi i}{10}} j\big)$\rule{0ex}{13pt} and golden ratio
$\tau=\frac12(1+\sqrt5)$ \\ with dot product $u\cdot
v={\textstyle\frac{1}{2}} (u\overline{v}+\overline{u}v)$ \rule{0ex}{13pt} and complex
conjugate $\overline{u_{1} +u_{2} i}=u_{1} -u_{2} i$.}}}
\end{center}
\caption{Quasicrystal construction using a continuous
quasicrystal phase function. Observe that a~quasicrystal point set
is contained within a level set of a continuous phase function
$f:{\mathbb C} \to {\mathbb R} $  def\/ined in the complex
plane~\cite{Dynamical generation of quasicrystals}. This
continuous phase function is formulated using the roots of the
non-crystallographic Coxeter group $H_2$, which comprise the
vertices  $\zeta ^{j} $  of a regular decagon. To generate a~quasicrystal point set, start by placing the f\/irst point at the
origin. For each newly placed point  $x\in {\mathbb C} $, consider
the candidate points  $z=x+\zeta ^{j} $  with  $j\in \{ 0,\ldots
,9\} $, which are the vertices of a regular decagon centered at
$x$, and only accept the candidates for which the phase function
$f(z)\ge T$  exceeds a desired threshold  $T$  that controls the
density of the resulting point set.}\label{Fig2}
\end{figure}

\begin{figure}[p]
\includegraphics[scale=0.98]{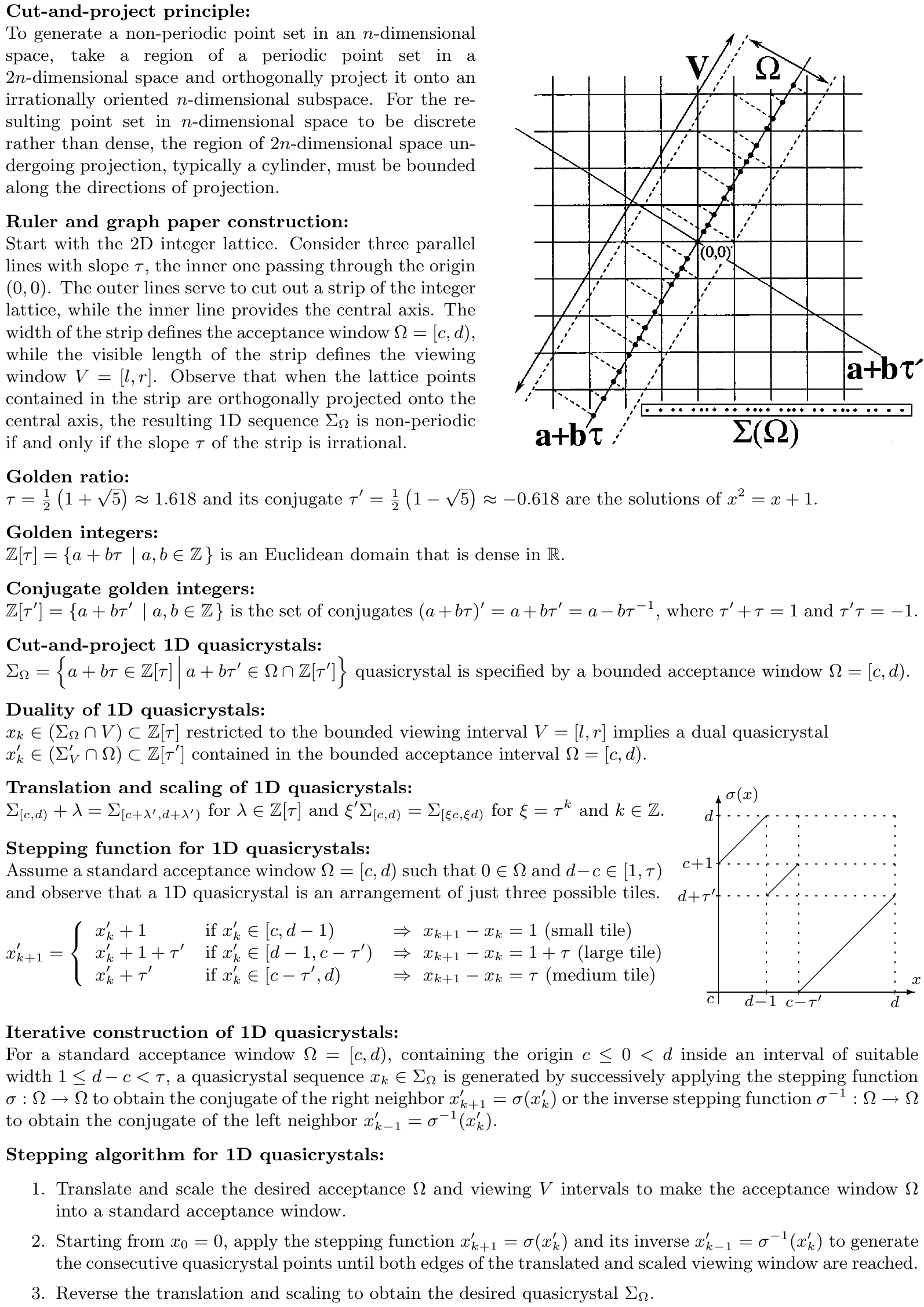}
\caption{Algorithm for 1D cut-and-project quasicrystals.}\label{Fig3}
\end{figure}

\begin{figure}[pt]
\includegraphics{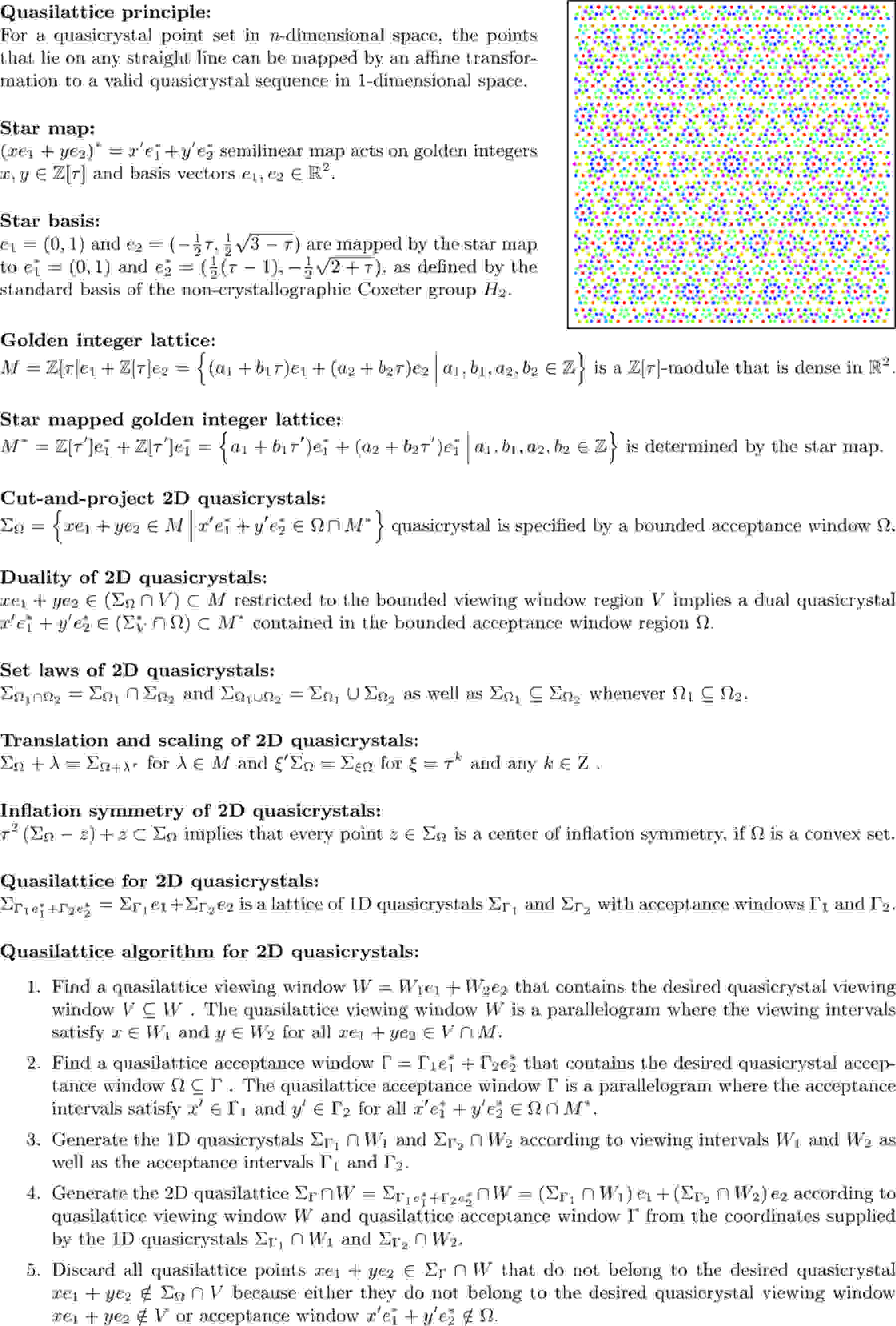}
\caption{Algorithm for 2D cut-and-project quasicrystals.}\label{Fig4}
\end{figure}

\begin{figure}[p]\small
\begin{center}
\includegraphics[width=1\textwidth]{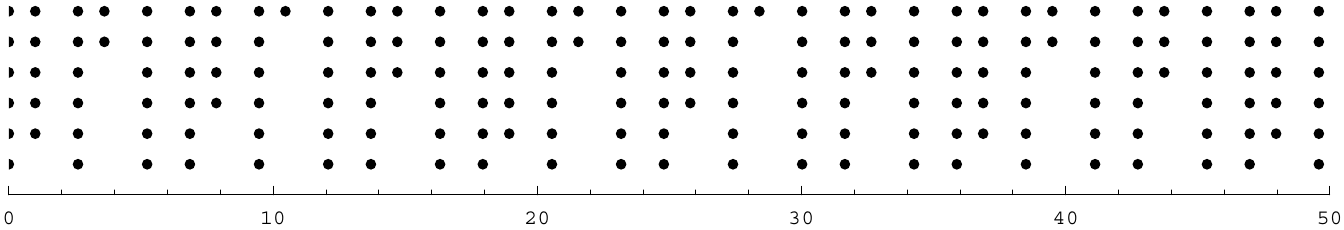}

{\footnotesize Evolution of a 1D quasicrystal point set
$\Sigma_{\Omega}$ from $\Omega =[0,1)$  to  $\Omega =[0,\tau )$.}
\bigskip

\setlength{\tabcolsep}{1.5pt}
\begin{tabular}{ccc}
\includegraphics[width=52mm]{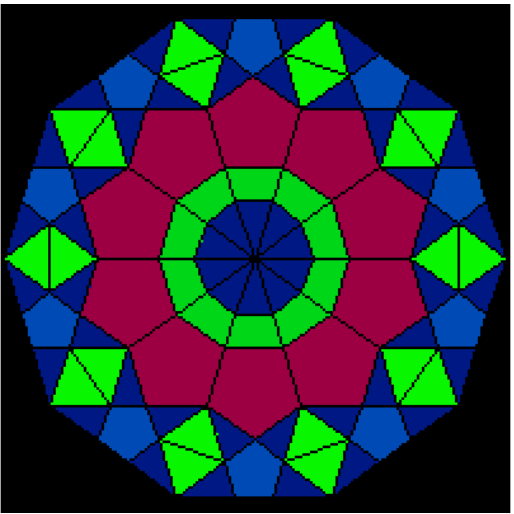}
&
\includegraphics[width=52mm]{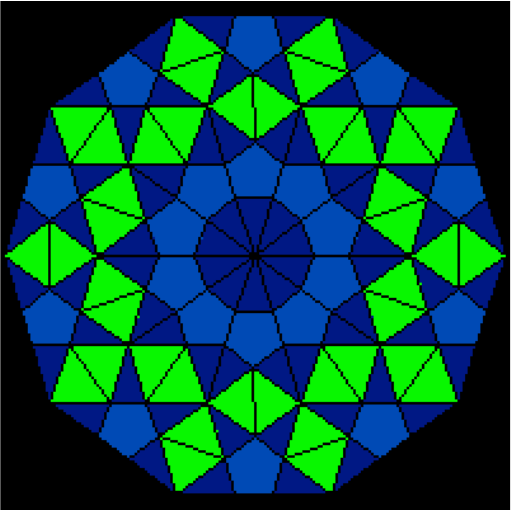}
&
\includegraphics[width=52mm]{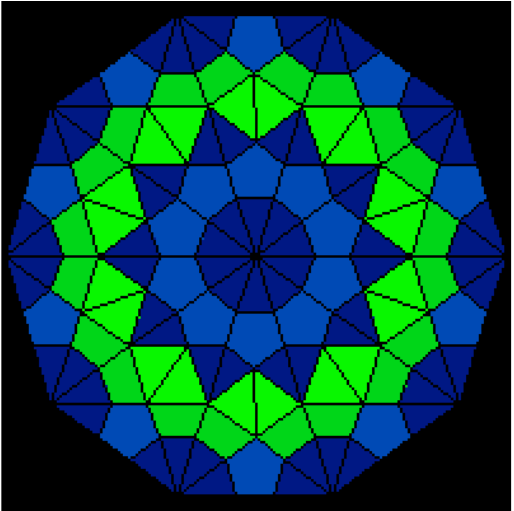}\\
\includegraphics[width=52mm]{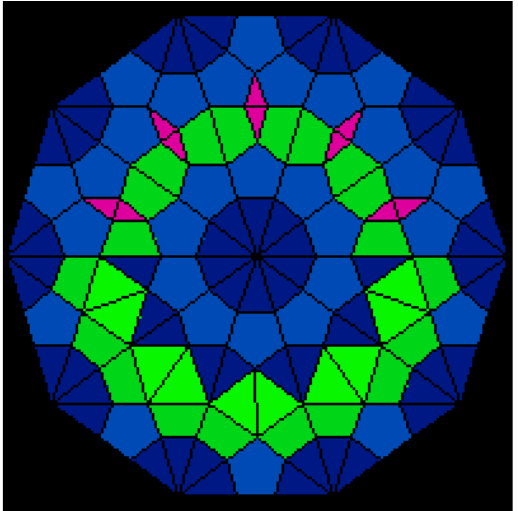}
&
\includegraphics[width=52mm]{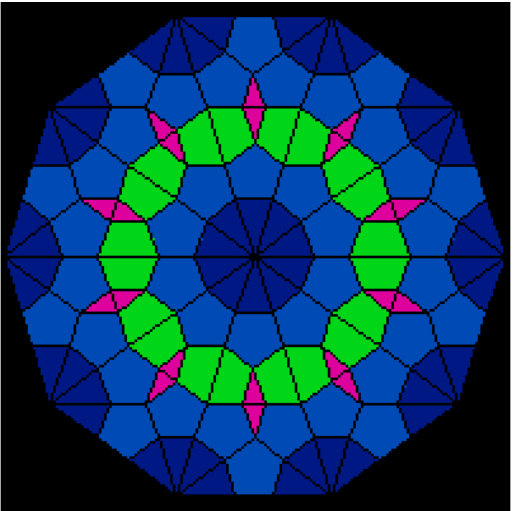}
&
\includegraphics[width=52mm]{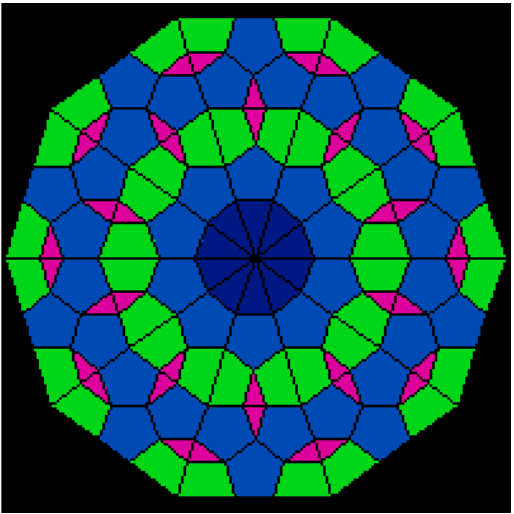}
\end{tabular}

{\footnotesize Evolution of a Delaunay graph of a 2D quasicrystal,
where 10 new sites are added at each generation.} \bigskip

\begin{tabular}{cc}
\includegraphics[width=70mm]{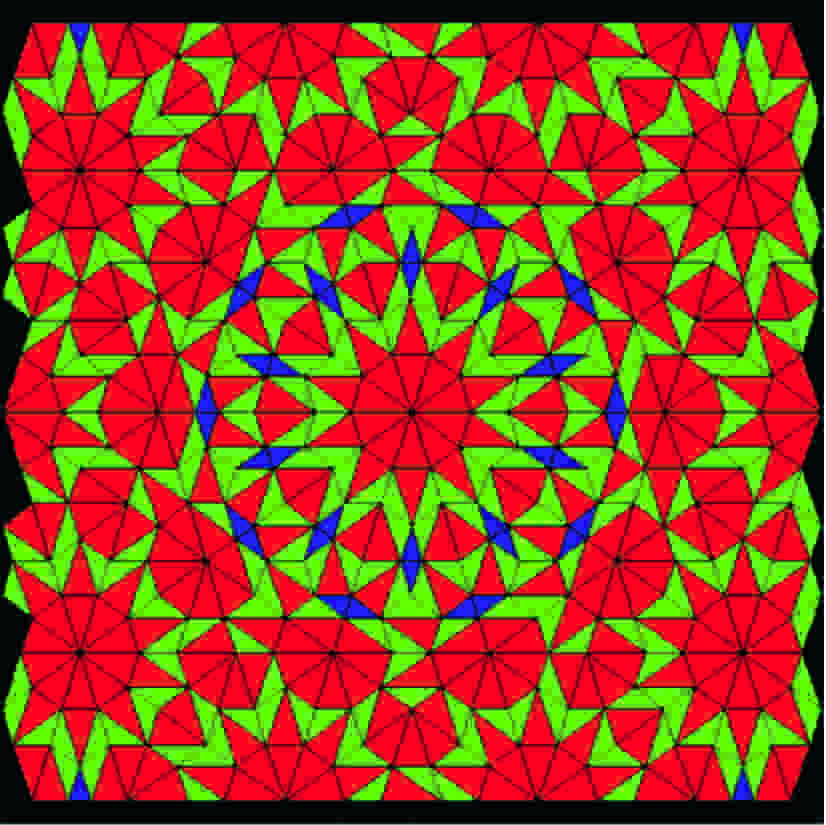}
 &
\includegraphics[width=70mm]{Fig1d-e}
\end{tabular}

{\footnotesize Evolution of a Delaunay triangulation of a 2D
quasicrystal, where the acceptance window is a decagon centered at
the origin that is expanded by a factor of $\tau$, from radius
$\tau^4+\tau^2$ on the left and to radius $\tau^5+\tau^3$ on the
right.}

\caption{Evolution of cut-and-project quasicrystals reveals their
self-similar structure.}\label{Fig5}
\end{center}
\end{figure}

We focus on a more general class of non-periodic point sets
corresponding to cut-and-project quasicrystals
\cite{Non-crystallographic root systems,Non-crystallographic root
systems and quasicrystals,Quasicrystals and geometry}. The
cut-and-project method was originally introduced by
Meyer~\cite{Algebraic numbers and harmonic analysis} in the
context of harmonic analysis and it was later adapted for
generating quasicrystals by Moody and Patera~\cite{Quasicrystals
and icosians}. The Fibonacci chain and Penrose tilings can be
regarded as special cases of such quasicrystals. In our work, we
employ the standard root lattice of the non-crystallographic
Coxeter group $H_2$, a group of ref\/lections taken from Lie algebra
theory. This approach to quasicrystals can be used to relate
discrete, non-periodic point sets and tilings (Fig.~\ref{Fig1}) with the
level sets of continuous, non-periodic functions (Fig.~\ref{Fig2}). To
produce a 2D cut-and-project quasicrystal, a 4D periodic lattice
is projected on a suitable 2D plane that is irrationally oriented
with respect to the lattice. Using this method, we obtain a dense
subspace consisting of integer coef\/f\/icient linear combinations of
the vertices of a regular decagon centered on the origin, which
are the roots of the non-crystallographic Coxeter group $H_2$.
This construction ensures that the coordinates of all its elements
can be expressed using only integers and an irrational number, the
golden ratio. To obtain a f\/inite 2D quasicrystal, we select only
those elements of the dense subspace contained in a specif\/ied
bounded region, called the viewing window, that are mapped by an
everywhere discontinuous algebraic transformation, called the star
map, to another specif\/ied bounded region, called the acceptance
window. Through the gradual expansion of a rotationally symmetric
acceptance window, a quasicrystal sequence can be uniquely ordered
by radial distance and angle from then center of the acceptance
window in order to produce a uniformly space-f\/illing point set in
the viewing window. As an important practical consequence, this
property directly enables progressive sampling. It could also
potentially enable adaptive sampling by varying the radius of the
acceptance window according to an application dependent importance
map def\/ined for the viewing window. A 2D quasicrystal can also be
expressed as a subset of a Cartesian product of two 1D
quasicrystals, in accordance with the fact that the points that
lie on any straight line through a 2D quasicrystal correspond to
some linearly transformed 1D quasicrystal. Therefore, in practice,
a 2D quasicrystal (Fig.~\ref{Fig4}) can be generated from a 2D lattice of
1D quasicrystals. Meanwhile, a 1D quasicrystal (Fig.~\ref{Fig3}) is
produced by taking a strip of a 2D periodic lattice, having f\/inite
width and irrational slope, and orthogonally projecting its points
onto a line of the same slope. The resulting 1D quasicrystal is
composed of at most three distinct tiles. It is easy to generate
1D quasicrystal points using an iterative numerical algorithm.
Alternatively, it is possible to exploit the self-similar
structure of a 1D quasicrystal, viewing it as the f\/ixed point of a
set of substitution rules that act recursively on a f\/inite
alphabet of possible tile arrangements.

Based on the geometry and algebra of the golden ratio, these
quasicrystal point sets exhibit some remarkable properties. They
display pentagonal and decagonal rotational symmetries, which
cannot occur in any periodic point set. Originally, the theory of
quasicrystals was motivated by solid state physics as a model of
the non-periodic geometric structures that describe the symmetries
of a new kind of long-range atomic order discovered in certain
metallic alloys~\cite{Schechtman}. While translational symmetries
def\/ine periodic crystals, inf\/lation symmetries can be used to
describe quasicrystals based on algebraic irrational numbers, such
as the golden ratio. When the acceptance window is a convex
region, every point of a quasicrystal can be viewed as a center of
inf\/lation symmetry. A quasicrystal can have no translational
symmetries and no periodic subsets. Moreover, it can be
partitioned into subsets such that each subset forms a valid
quasicrystal. Consider the local conf\/igurations of tiles in an
inf\/inite quasicrystal mosaic formed by a~Voronoi diagram or a
Delaunay triangulation. According to the repetitivity property
implied by suitable regularity conditions, each fragment is
repeated inf\/initely many times in the mosaic. Yet no single
fragment is ever suf\/f\/icient to determine the structure of the
whole mosaic because every f\/inite fragment, no matter its size,
occurs in an uncountable inf\/inity of nonequivalent mosaics.
Furthermore, according to the Delaunay point set property,
quasicrystals are both uniformly discrete and relatively dense,
creating space-f\/illing tilings.

As Delaunay point sets, quasicrystals are particularly well suited
to image sampling. They enforce both a minimal and a maximal
distance between each sample site and its closest neighboring
site. In quasicrystal sampling, we rely on the golden ratio to
ensure symmetry and self similarity, which are generally absent
when other than algebraic irrational numbers are used with the
cut-and-project method to produce non-periodic point sets. By
taking this approach, we can endow a rendition with a decorative
symmetry, which viewers may f\/ind attractive in the context of
non-photorealistic rendering. Compared with the regular grids of
periodic point sets, the self-similar, space-f\/illing structure of
non-periodic point sets (Fig.~\ref{Fig5}) appears less monotonous,
especially during progressive image rendering. In ef\/fect, the
geometric structure of quasicrystal sampling eliminates the
possibility of aliasing artifacts regularly repeating in the
rendition. However, in quasicrystal sampling, f\/ixed local
conf\/igurations of sample sites can be repeated at multiple places
and orientations, albeit not at regular intervals, with the
potential to yield some recurring, anisotropic aliasing artifacts.
Although we did do so in this work, for photorealistic image
reconstruction, it is preferable to avoid inducing global
rotational symmetry in the sampling pattern, which is done by
ensuring the viewing window does not contain the origin when the
acceptance window is symmetric with respect to the origin.

\section{Evaluation}

We now present a qualitative evaluation of quasicrystal sampling
in the context of non-adaptive sampling strategies for use in
image representation. In this application, non-adaptive sampling
strategies serve as building blocks for interactive sampling,
adaptive sampling, and importance sampling techniques. In ef\/fect,
they dictate the placement of sample sites in image regions
sampled at a uniform resolution. For this purpose, a non-adaptive
sampling strategy should satisfy several image representation
objectives. The sample sites should be distributed in a manner
that fairly and accurately represents the image. The sampling
pattern should be evenly space-f\/illing in order to enable
progressive image rendering. Without any preconceptions about the
distribution of visually salient features in the image, the same
amount of information should be devoted to capturing each part of
the image. Hence, the number of sample sites placed in any region
of the image should be proportional to its area, so the sampling
density remains the same throughout the image. Both globally and
locally, the placement of sample sites should be uniform and
isotropic while still allowing for a variety of dif\/ferent sample
site conf\/igurations. To prevent aliasing, the sample sites should
not be arranged into f\/ixed conf\/igurations that visibly repeat
locally or globally. To prevent clustering, a minimum distance
between sample sites should be maintained throughout the image.
Assuming that the correlation between pixels decreases with
distance, a sample site should be placed close to the centroid of
its Voronoi polygon to enable its sampled color to be most
representative of its region of inf\/luence when the image is
reconstructed using a local interpolation technique. Naturally,
the relative importance of these considerations depends on the
requirements of a particular application. We have formulated these
priorities based on our previous work on a multiresolution image
representation~\cite{Stylized multiresolution image
representation} designed to simultaneously support both
photorealistic image reconstruction and non-photorealistic image
rendering. While previous qualitative surveys~\cite{Principles of
digital image synthesis} and quantitative comparisons
\cite{Discrepancy as a quality measure for sample distributions}
of image sampling have focused on applications in photorealistic
image reconstruction, they did not cover quasicrystal sampling and
farthest point sampling. Furthermore, their tests were not carried
out on an image representation of digital photographs and they did
not specif\/ically address the needs of non-photorealistic image
rendering.

We compared quasicrystal sampling to a number of standard
non-adaptive image sampling strategies~\cite{Principles of digital
image synthesis}, ref\/lecting dif\/ferent approaches to the inherent
trade-of\/f between aliasing and noise. For our evaluation, we chose
approaches that exemplify divergent aims in sampling. For each
approach, we selected a representative implementation. As noted
below, alternative implementations are certainly possible but they
are likely to produce similar qualitative results. From the
deterministic to the stochastic, we tested a range of sampling
strategies (Fig.~\ref{Fig7}):

\begin{enumerate}\itemsep=0pt

\item {\bf  Periodic sampling}~\cite{Mathematical tools for
computer-generated ornamental patterns} aims for global
regularity. Our implementation relies on a square lattice ref\/ined
in scan line order. An alternative implementation could use a
hexagonal lattice, the densest periodic lattice in the plane.

\item {\bf  Quasicrystal sampling}~\cite{Quasicrystals and
geometry} aims for local regularity. Our implementation relies on
the cut-and-project method applied using the golden ratio. An
alternative implementation could use a Penrose tiling produced
using a hierarchical substitution algorithm.

\item {\bf  Farthest point sampling}~\cite{FarthestPointStrategy}
aims for spatial uniformity. Our implementation relies on the
principle of progressively sampling at the point of least
information, placing each new sample site at the point farthest
from any preceding sample site, which is necessarily a vertex of
the Voronoi diagram of the preceding sample sites. An alternative
implementation could position sample sites to conform to a
centroidal Voronoi diagram so that each sample site is placed at
the centroid of its Voronoi polygon.

\item {\bf Jittered sampling}~\cite{Filtered jitter} aims for
local variability. Our implementation relies on a full random
displacement of a square lattice ref\/ined in scan line order. An
alternative implementation could use a partial random displacement
of a hexagonal lattice.

\item {\bf  Quasirandom sampling}~\cite{Numerical recipes in c}
aims for low discrepancy. Our implementation relies on the Halton
sequence. An alternative implementation could use a Sobol
sequence.

\item {\bf Random sampling}~\cite{Principles of digital image
synthesis} aims for global variability. Our implementation relies
on a uniform distribution. An alternative implementation could use
a random walk on a unit square with toroidal boundary conditions.
\end{enumerate}

To perform a qualitative evaluation of the image sampling
strategies, we applied a number of computer visualization
techniques. For each non-adaptive sampling strategy, we visualize
its sample sites (Fig.~\ref{Fig7}) in the spatial domain using a Voronoi
diagram (Fig.~\ref{Fig8}) and in the frequency domain using a Fourier
power spectrum (Fig.~\ref{Fig9}). We examined the visual ef\/fects of
applying the image sampling strategies in the context of various
image rendering techniques that are used in multiresolution image
representations. For photorealistic image reconstruction~\cite{ScatteredData}, we tested the accuracy of Shepard's
interpolation (Fig.~\ref{Fig13}), an inverse distance weighted
interpolation method that applies the Voronoi diagram to determine
the color of each pixel based on its four nearest sample sites, as
well as Gouraud shading (Fig.~\ref{Fig14}), a piecewise linear
interpolation method that applies the Delaunay triangulation to
determine the color of each pixel based on its three surrounding
sample sites. To quantitatively assess the results of these widely
used local interpolation techniques, we relied on the peak
signal-to-noise ratio (PSNR). This standard image f\/idelity metric
\cite{Digital color imaging handbook} estimates the accuracy of
the rendition according to the negative logarithm of the mean
squared error between the rendered and actual RGB color values of
the pixels. Hence, higher peak signal-to-noise ratio scores are
considered better. For non-photorealistic image rendering
\cite{Stylized multiresolution image representation}, we
experimented with a simple ``paint strokes'' rendering style
(Fig.~\ref{Fig12}), which applies geometric subdivision to the Delaunay
triangulation of the sample sites and then uses linear and
nonlinear interpolation to emphasize the transitions between the
sampled colors.

\begin{figure}[t]
\centering
\setlength{\tabcolsep}{2pt}
\begin{tabular}{cccc}
\includegraphics[scale=0.35]{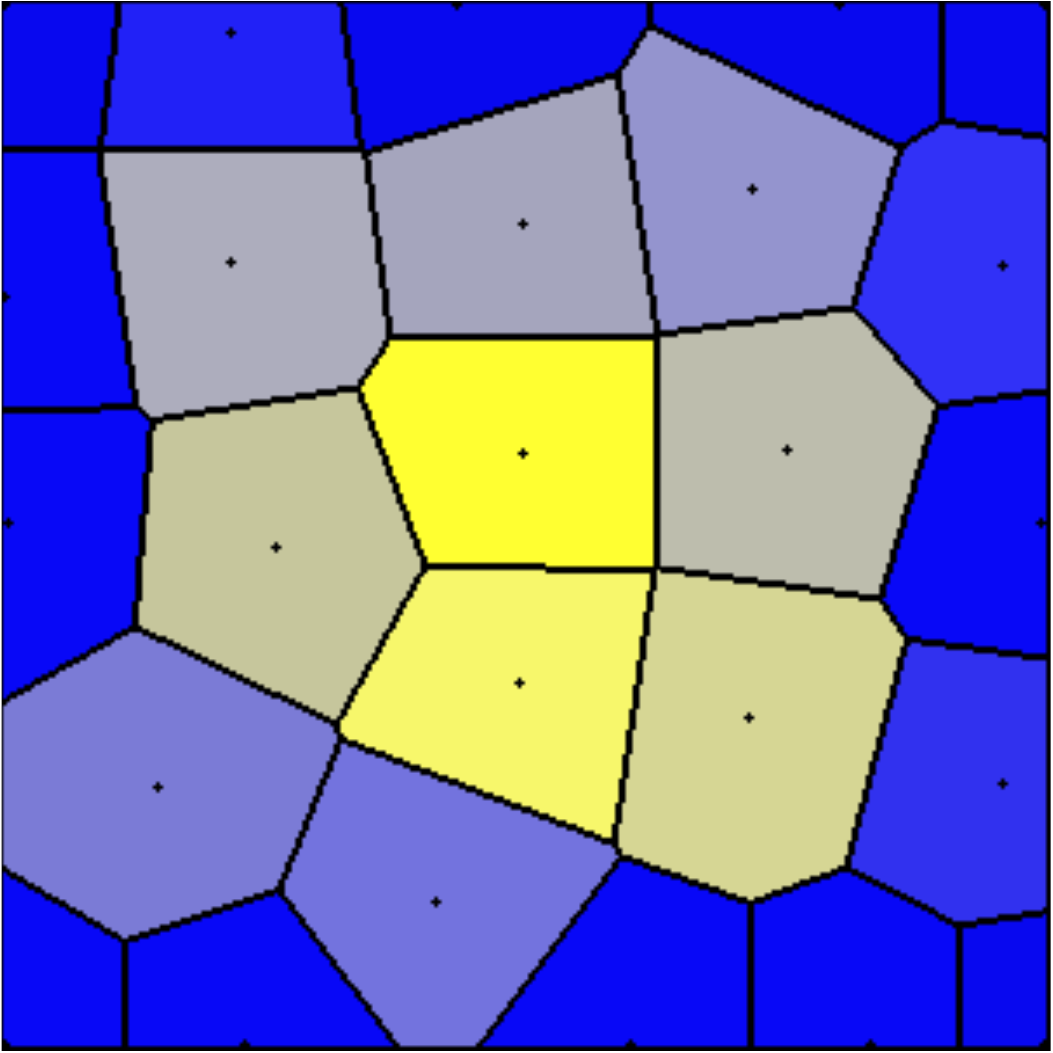}
&
\includegraphics[scale=0.35]{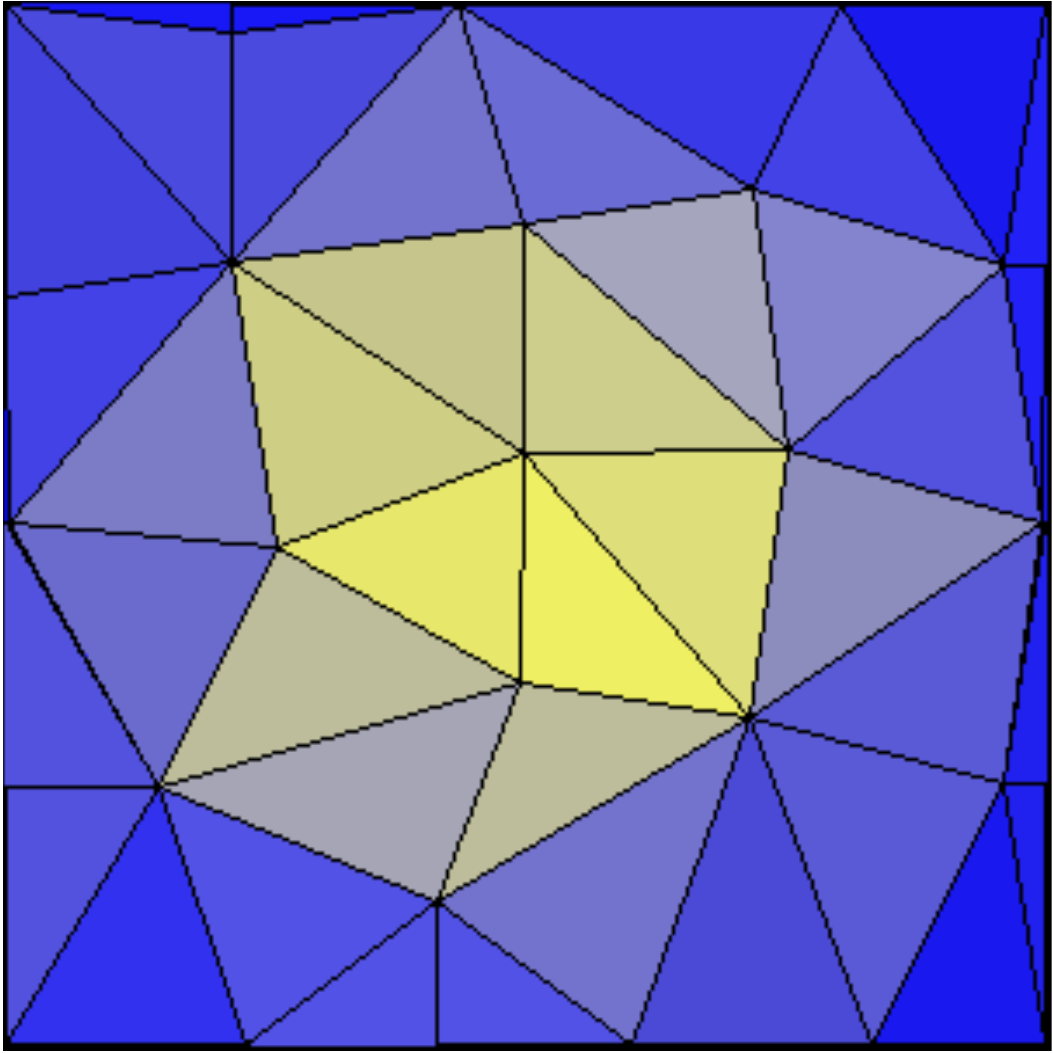}
&
\includegraphics[scale=0.35]{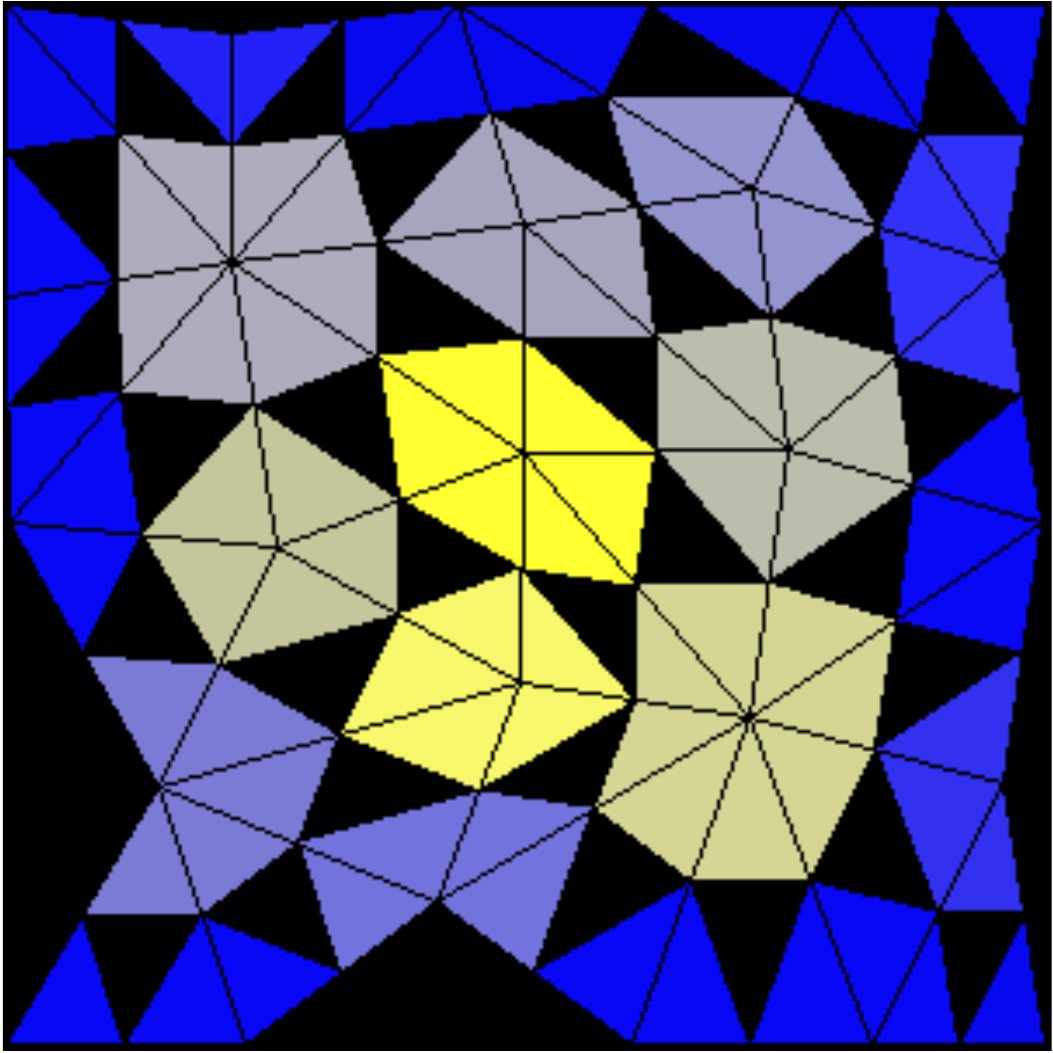}
&
\includegraphics[scale=0.35]{Fig6c-e}
\end{tabular}
\caption{Techniques for visualizing the spatial conf\/iguration of
sample sites: Voronoi diagram (far left), Delaunay triangulation
(center left), Delaunay star shapes (center right), and our mosaic
rendering technique (far right).}\label{Fig6}
\vspace{-2mm}
\end{figure}

Our mosaic rendering style (Fig.~\ref{Fig6}) of\/fers a new computer
visualization tool for evaluating the spatial properties of point
sets, such as the sample sites produced by the various image
sampling strategies. To produce a mosaic rendering, we apply
geometric subdivision to the Delaunay triangulation of the sample
sites. The midpoints of the edges of each Delaunay triangle are
joined to form three outer triangles and one inner triangle. Each
outer triangle is rendered with the color sampled at its vertex of
the original Delaunay triangle, while the central triangle is
colored black. As each sample site is represented by a star-shaped
polygonal tile, the resulting mosaic appears packed as tightly as
possible, with the black central triangles serving as grout
between the tiles. As a sample site's local neighborhood
(Fig.~\ref{Fig6}, far left) comprises the surrounding sample sites
connected to it by edges in the Delaunay triangulation (Fig.~\ref{Fig6},
center left), the sample site's mosaic tile (Fig.~\ref{Fig6}, far right)
is shaped to ref\/lect the star of its surrounding Delaunay
triangles (Fig.~\ref{Fig6}, center right). For instance, a sample site's
mosaic tile is a convex or concave polygon according to whether
its neighboring sites are arranged in a convex or concave
conf\/iguration. As a visualization tool, the advantage of mosaic
rendering is that the layout of the star-shaped mosaic tiles makes
the spatial properties of a sampling strategy easier to see at a
glance than the triangles of a Delaunay triangulation or the
convex polygons of a Voronoi diagram. The sizes of the mosaic
tiles are indicative of the uniformity of sampling, as coarsely
sampled regions give rise to large tiles and f\/inely sampled
regions give rise to small tiles, making common defects such as
clustering, undersampling, and oversampling easy to detect. The
orientations of the mosaic tiles are indicative of the isotropy of
sampling, as the preferred directions of the sampling are revealed
in the preferred rotations of the tiles, making global or local
grid structures easy to detect. The shapes of the mosaic tiles are
indicative of the heterogeneity of sampling, as the local
conf\/igurations of neighboring sites uniquely determine the tile
polygons, making repetitive patterns easy to detect. For instance,
farthest point sampling produces tiles of uniform size and similar
shape to create the appearance of a pebble mosaic, while
quasicrystal sampling yields a~decorative tiling with just a small
set of possible tile shapes.

Our qualitative evaluation of image sampling strategies uses seven
criteria (Fig.~\ref{Fig10}) known to af\/fect the visual quality of
photorealistic image reconstruction and non-photorealistic image
rendering~\cite{Stylized multiresolution image representation}:

\begin{enumerate}\itemsep=0pt
\item{\bf Accurate reconstruction} requires the rendition to
faithfully represent the likeness of the original image. It is a
necessary but not suf\/f\/icient condition of success in both
photorealistic and non-photorealistic image rendering. This
objective appears to be closely associated with uniform coverage
and centroidal regions. It is assessed by measuring the peak
signal-to-noise ratio for the results of photorealistic image
reconstruction (Figs.~\ref{Fig13} and~\ref{Fig14}). Its most pronounced
ef\/fects can also be observed in the results of non-photorealistic
image rendering (Fig.~\ref{Fig12}). When the resolution of sampling is
uniform, periodic sampling yields the most accurate image
interpolation (in Fig.~\ref{Fig14}, for the regular square grid, this
takes place when there are $33^2=1089$, $65^2=4225$, and
$129^2=16641$ samples). However, when regions of varying
resolution arise during progressive ref\/inement, the accuracy of
periodic sampling can substantially deteriorate. Similar behavior
is observed in jittered sampling since it applies random
perturbations to a periodic point set. Farthest point sampling
produces image reconstructions that are nearly as accurate as
periodic sampling, but its performance does not diminish during
progressive ref\/inement. Intermediate accuracy is of\/fered by
quasicrystal sampling, which appears to be slightly more accurate
than quasirandom sampling. The least accurate reconstructions are
produced by jittered and random sampling. Given its popularity in
computer graphics implementations, the poor performance of
jittered sampling is rather disappointing. Of course, the accuracy
of jittered sampling can always be made closer to that of periodic
sampling by reducing the amount of random displacement, which
risks reintroducing the aliasing artifacts of periodic sampling.
In general, accuracy is improved by uniformity and reduced by
randomness, an ef\/fect that can be readily seen as producing tight
or loose image stylization.

\begin{figure}[t]
\centering
\setlength{\tabcolsep}{2pt}
\begin{tabular}{ccc}
\includegraphics[scale=0.59]{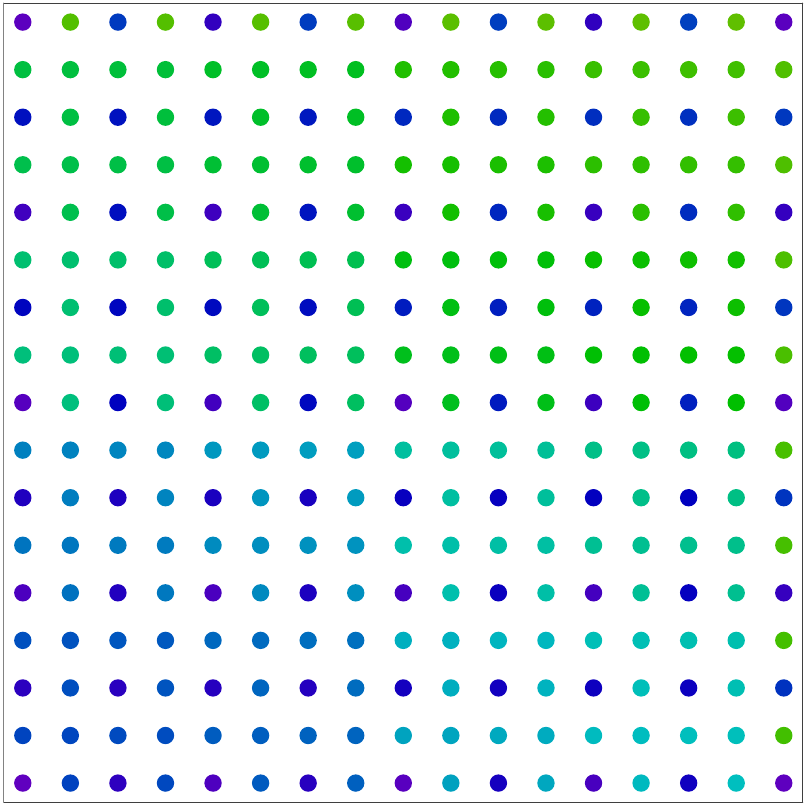}
&
\includegraphics[scale=0.59]{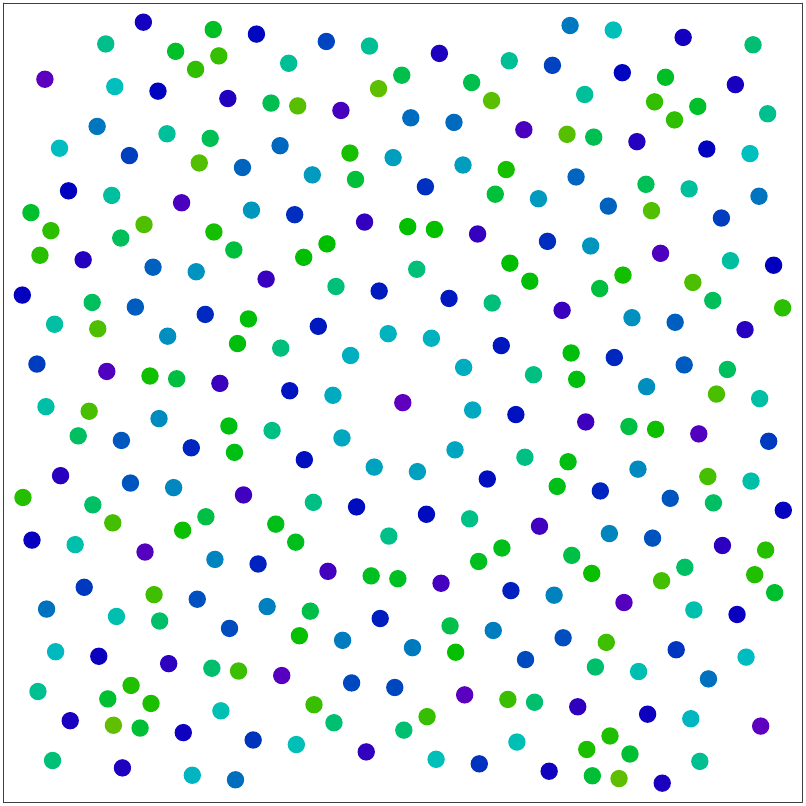}
&
\includegraphics[scale=0.59]{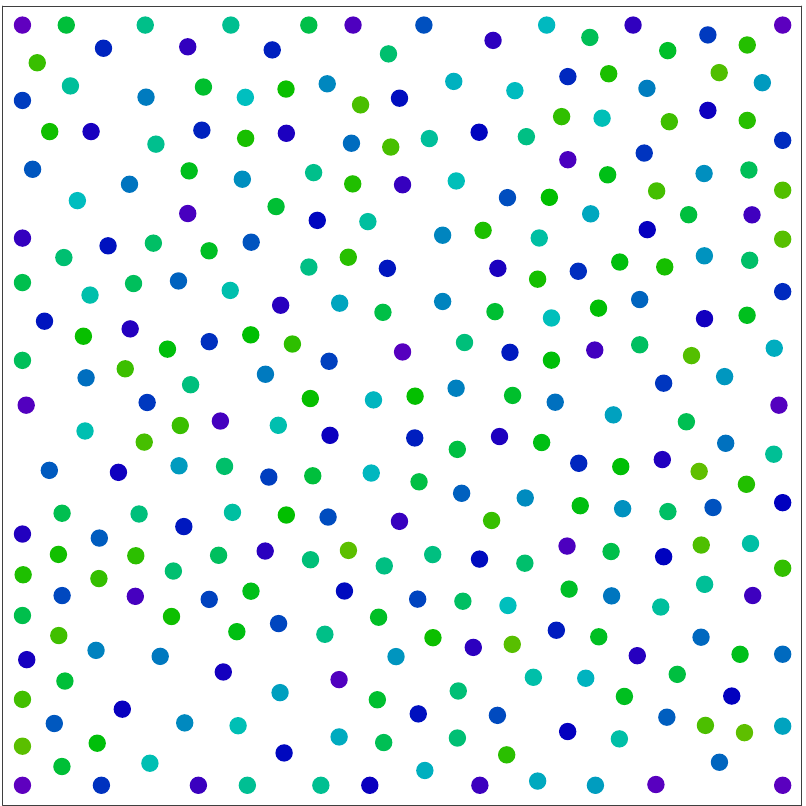}
\\
\includegraphics[scale=0.59]{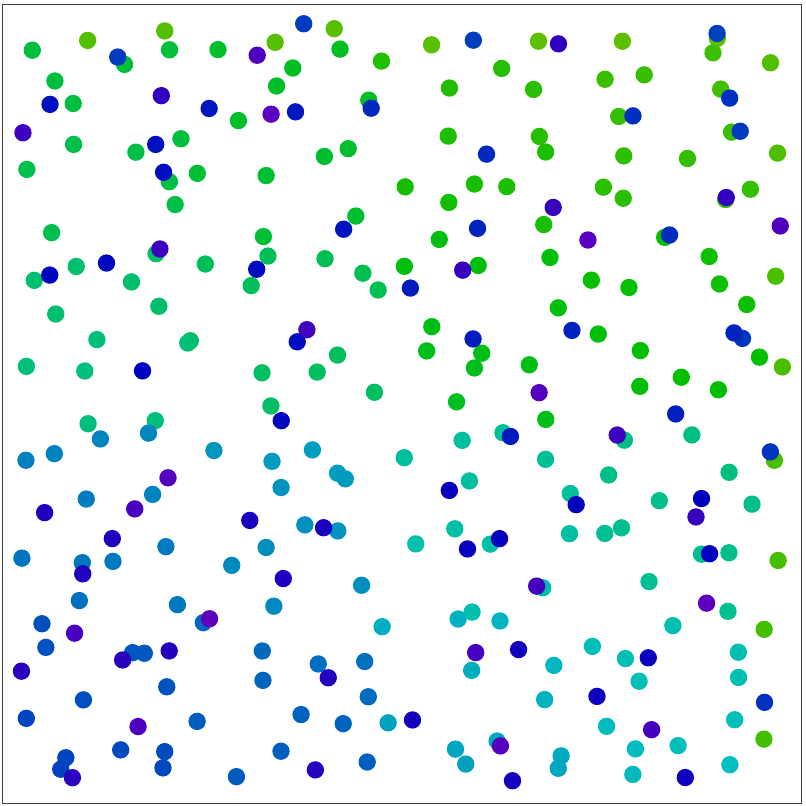}
&
\includegraphics[scale=0.59]{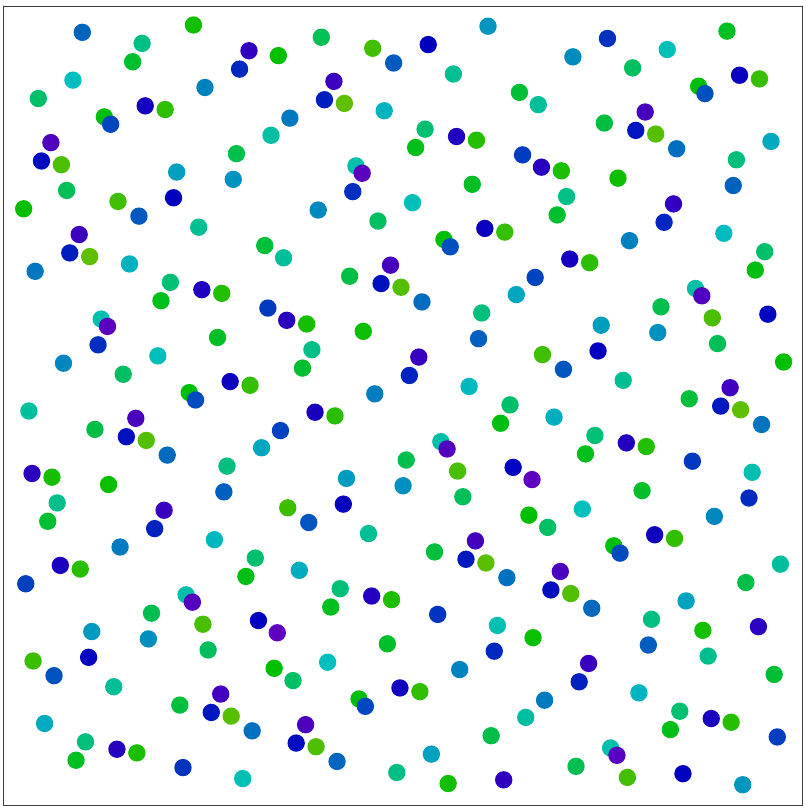}
&
\includegraphics[scale=0.59]{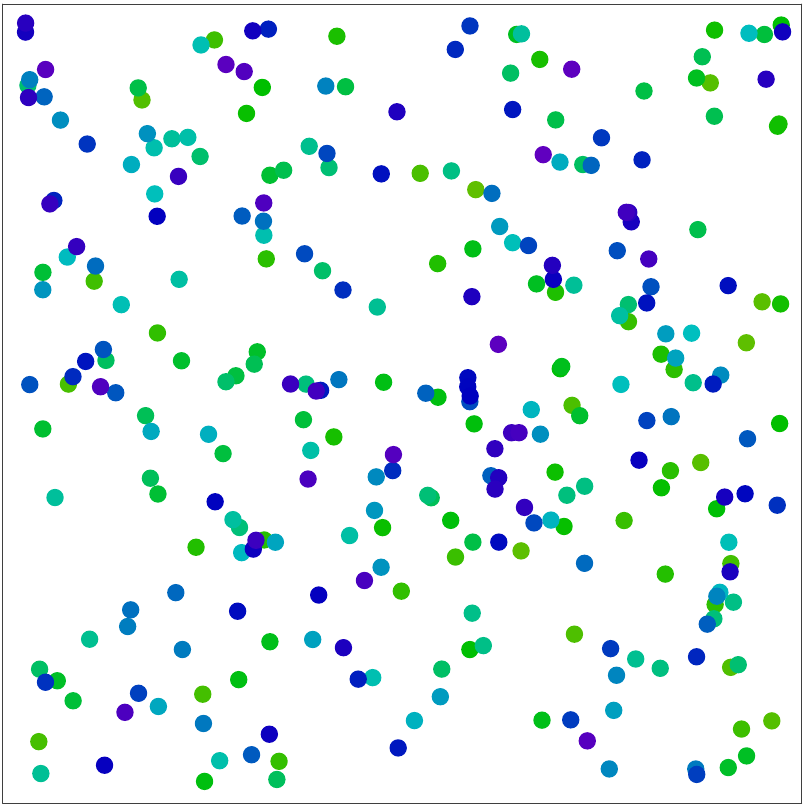}
\end{tabular}
\caption{Non-adaptive sampling strategies: periodic sampling (top
left), quasicrystal sampling (top center), farthest point sampling
(top right), jittered sampling (bottom left), quasirandom sampling
(bottom center), and random sampling (bottom right). Sampling
starts with the dark blue sites and f\/inishes with the light green
sites. }\label{Fig7}\vspace{-2mm}
\end{figure}

\item{\bf Progressive ref\/inement} requires the sample sites to
smoothly f\/ill the available space, avoiding abrupt changes in
appearance as new sample sites are sequentially added to the
rendition. This objective serves to enable a multiresolution image
representation to support progressive rendering of compressed
images based on an incremental sampling of the image data. It is
assessed by examining the spatial layout of the sequence of sample
sites (Fig.~\ref{Fig7}). Under ideal circumstances, progressive
ref\/inement should yield a smooth curve for the peak
signal-to-noise ratio (Fig.~\ref{Fig14}). The best progressive ref\/inement
results are produced by farthest point sampling and quasicrystal
sampling, as these methods maintain a uniform sampling density by
ensuring that new sample sites are placed in the largest empty
spaces between the existing sample sites. Quasirandom sampling
proves slightly less prof\/icient, as it places some sample sites
very close together while keeping others far apart. Random
sampling is even less ef\/fective due to its tendency to locally
cluster sample sites. The regular grids used in periodic and
jittered sampling are not suitable for smooth progressive
ref\/inement, especially when they are ref\/ined in scan line order.
While other ref\/inement schemes can be applied to regular grids,
such as ref\/inement in random order, their intrinsic symmetry makes
it dif\/f\/icult to smoothly increase the sampling density throughout
the image.

\item{\bf Uniform coverage} \looseness=1 requires the sample sites to be evenly
distributed regardless of position, avoiding conf\/igurations that
place sample sites too close or too far from their nearest
neighbors. This objective is assessed using Voronoi diagrams
(Fig.~\ref{Fig8}) as well as mosaic renderings (Fig.~\ref{Fig11}). Its ef\/fect
determines the sizes of brush strokes in non-photorealistic image
rendering (Fig.~\ref{Fig12}). Uniform coverage is associated with a
Fourier power spectrum (Fig.~\ref{Fig9}) that displays an empty ring
around the central spike, as low frequencies are attenuated in
favor of a threshold frequency corresponding to the most commonly
observed nearest neighbor distance between sample sites. Although
a blue noise spectrum can ensure uniform coverage, it is not a
necessary condition. When the resolution of sampling is uniform,
periodic sampling generates uniform coverage, as its mosaic tiles
are all exactly the same size. However, periodic sampling cannot
sustain uniform coverage during progressive ref\/inement. By design,
farthest point sampling maintains uniform coverage at all times,
as its mosaic tiles are all approximately the same size.
Quasicrystal sampling maintains nearly as uniform coverage, as its
mosaic tiles are limited to just a few comparable sizes. While
quasirandom and jittered sampling strive to uphold a uniform
density of sampling, they nevertheless are less ef\/fective at
providing uniform coverage, as their mosaic tiles come in many
sizes. In the case of jittered sampling, uniform coverage can be
improved by reducing the amount of random displacement. Random
sampling does not give uniform coverage, as its mosaic tiles
exhibit the greatest range of dif\/ferent sizes.

\begin{figure}[ht]
\centering
\includegraphics[width=152mm]{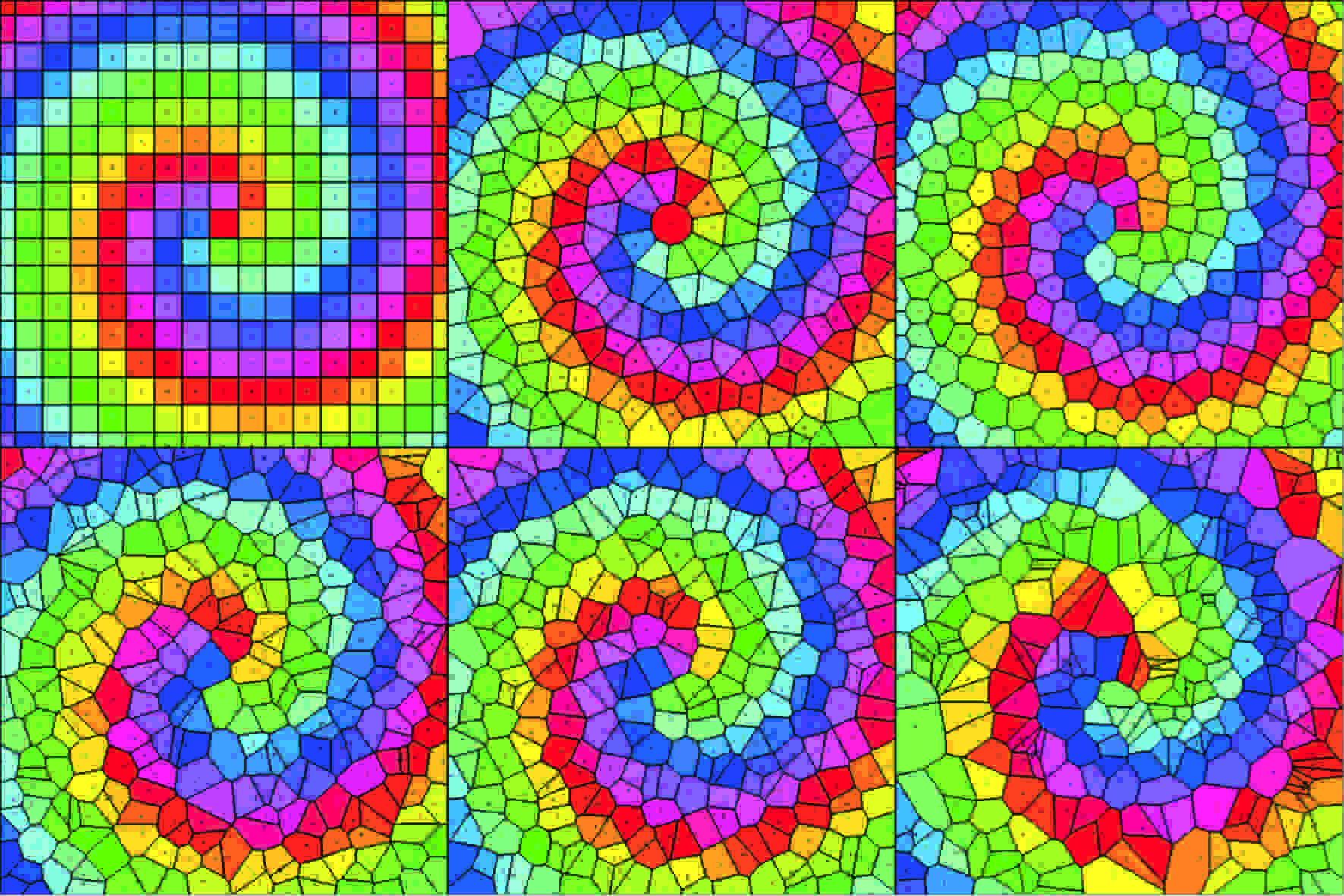}

\caption{Voronoi diagrams of image sampling strategies applied to
a color spiral test image: periodic sampling (top left),
quasicrystal sampling (top center), farthest point sampling (top
right), jittered sampling (bottom left), quasirandom sampling
(bottom center), and random sampling (bottom right). }\label{Fig8}\vspace{-2mm}
\end{figure}

\item {\bf Isotropic distribution}  requires the sample sites to be
evenly distributed regardless of orientation, avoiding
conf\/igurations that align sample sites along globally or locally
preferred directions. This objective is assessed using Voronoi
diagrams (Fig.~\ref{Fig8}) as well as mosaic renderings (Fig.~\ref{Fig11}). Its
ef\/fect is to determine the orientations of brush strokes in
non-photorealistic image rendering (Fig.~\ref{Fig12}). An isotropic
distribution produces a Fourier power spectrum (Fig.~\ref{Fig9}) that
displays a rotational symmetry around the central spike, as the
power at each frequency does not depend on its orientation.
Although a blue noise spectrum can ensure isotropic distribution,
it is not a necessary condition. Random sampling is the most
isotropic, as its sample sites are both locally and globally
uncorrelated. Farthest point and jittered sampling are nearly as
isotropic, as their sample sites can exhibit slight local
alignment. Farthest point samples can appear to be placed in
roughly hexagonal local conf\/igurations. Jittered samples can
appear to retain some of the structure of the underlying square
grid, as the isotropy of jittered sampling ref\/lects the amount of
random displacement used to generate the sampling. Quasirandom
sampling has intermediate isotropy, as its sample sites can
exhibit slight global alignment, which can be verif\/ied in the lack
of radial symmetry in its Fourier power spectrum. Periodic
sampling and quasicrystal sampling do not have isotropic
distribution since their sample sites are globally aligned along
predetermined axes.

\begin{figure}[ht]
\centering
\setlength{\tabcolsep}{2pt}
\begin{tabular}{ccc}
\includegraphics[width=46.5mm]{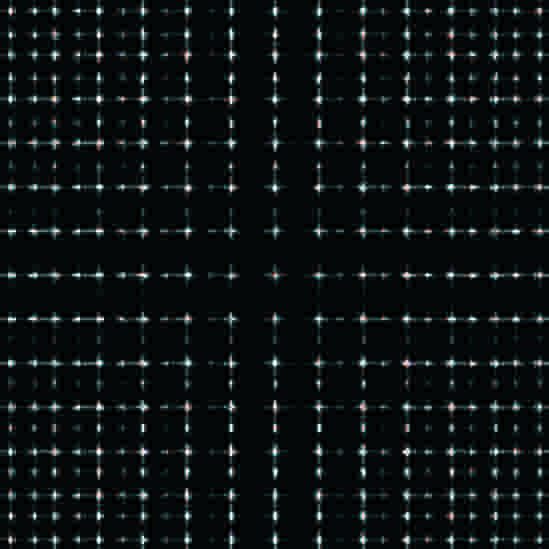}
&
\includegraphics[width=46.5mm]{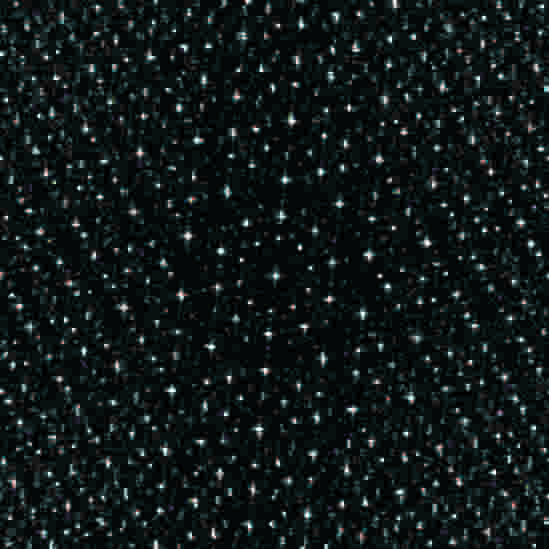}
&
\includegraphics[width=46.5mm]{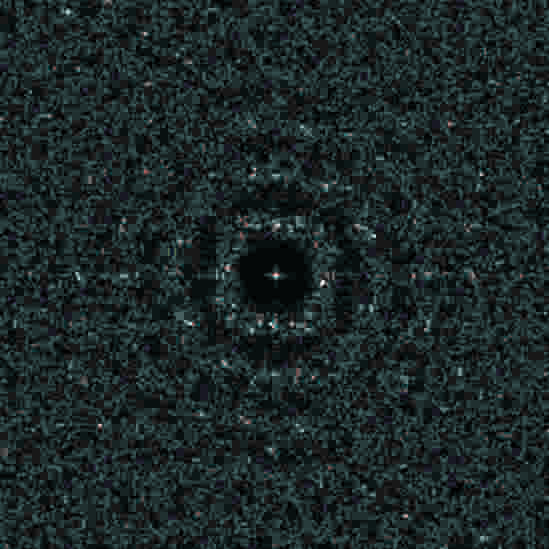}
\\
\includegraphics[width=46.5mm]{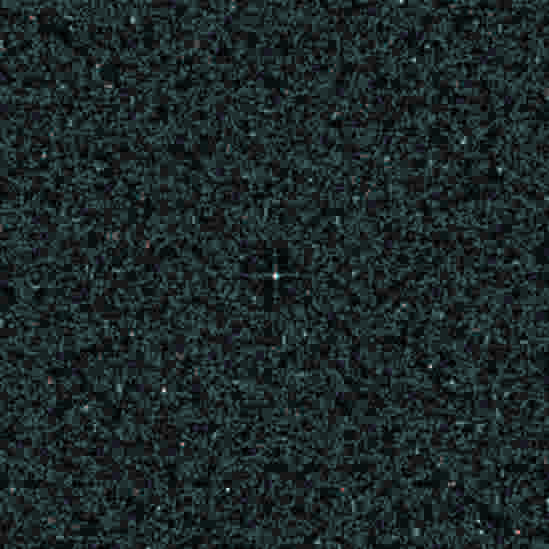}
&
\includegraphics[width=46.5mm]{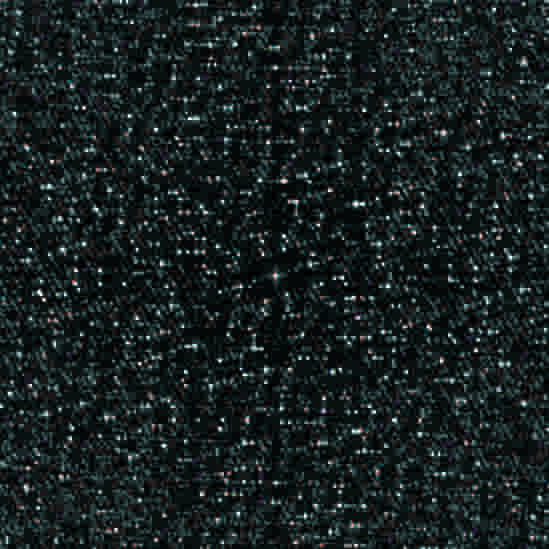}
&
\includegraphics[width=46.5mm]{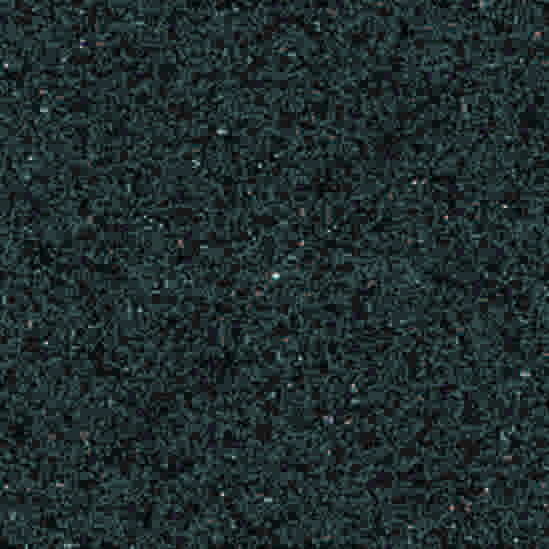}
\end{tabular}
\caption{Fourier power spectrums of image sampling strategies:
periodic sampling (top left), quasicrystal sampling (top center),
farthest point sampling (top right), jittered sampling (bottom
left), quasirandom sampling (bottom center), and random sampling
(bottom right). }\label{Fig9}\vspace{-2mm}
\end{figure}

\item{\bf Blue noise spectrum} \looseness=-1 requires the sample sites to be
distributed similarly to a Poisson disk distribution, a random
point f\/ield conditioned on a minimum distance between the points.
According to this objective, for an image sampling strategy to
provide ef\/fective antialiasing for image rendering, it should
attempt to mimic the idealized distribution of photoreceptors in
the human eye. Usually implying both uniform coverage and
isotropic distribution, a blue noise spectrum is highly desirable
in many computer graphics applications, particularly
photorealistic image reconstruction. It is assessed by examining
the Fourier power spectrums of the sampling strategies (Fig.~\ref{Fig9})
for a radially symmetric prof\/ile that concentrates noise in the
high frequencies while attenuating the power of the low
frequencies, thereby eliminating the aliasing artifacts associated
with low frequency patterns that can appear distracting to the
eye. In ef\/fect, a blue noise spectrum exhibits a disk of low power
around the origin, surrounded by roughly constant power at the
higher frequencies. Its ef\/fects can be judged according to the
amount of aliasing present in photorealistic image reconstruction
(Fig.~\ref{Fig13}) and non-photorealistic image rendering (Fig.~\ref{Fig12}).
Farthest point sampling has a Fourier power spectrum that is
closest to a blue noise spectrum. Jittered sampling attempts to
replicate the blue noise spectrum, but it does not clearly exhibit
the threshold frequency ripple around the central spike.
Quasirandom is even less successful because its Fourier power
spectrum lacks radial symmetry. Periodic, quasicrystal, and random
sampling have Fourier power spectrums that do not resemble the
blue noise spectrum. The Fourier power spectrums of periodic and
quasicrystal sampling ref\/lect the spatial structures and
directional symmetries of the lattices used to place the sample
sites. On the other hand, the white noise spectrum of random
sampling assigns roughly the same power to all frequencies.

\item{\bf Centroidal regions} require sample sites to be well
centered with respect to their Voronoi polygons, approximating a
centroidal Voronoi diagram. Typically associated with uniform
coverage and accurate reconstruction, this objective is popular in
non-photorealistic image rendering. Sampling strategies, such as
periodic sampling, that produce centroidal regions can still be
prone to aliasing artifacts since centroidal regions do not
guarantee a blue noise spectrum. Centroidal regions can be readily
assessed using the Voronoi diagrams (Fig.~\ref{Fig8}). The ef\/fects can
also be observed in the shapes of tiles in mosaic rendering
(Fig.~\ref{Fig11}) and brush strokes in non-photorealistic image
rendering (Fig.~\ref{Fig12}). Periodic sampling, placing each sample site
at the same distance from all of its nearest neighbors, generates
exact centroidal Voronoi regions. Quasicrystal and farthest point
sampling produce approximately centroidal Voronoi regions. To
place sample sites close to the center of their Voronoi polygons,
quasicrystal sampling relies on local symmetries while farthest
point sampling relies on nearest neighbor distance. Jittered
sampling and quasirandom sampling have dif\/f\/iculty ensuring
centroidal regions because they are less ef\/fective at maintaining
a minimal nearest neighbor distance. Finally, random sampling can
only produce centroidal regions by chance.

\item{\bf Heterogeneous conf\/igurations} require sample sites to be
placed in a variety of dif\/ferent local arrangements, avoiding
regularly or randomly repeating the same sampling patterns. While
this objective is not traditionally a concern in photorealistic
image reconstruction, it helps to prevent non-photorealistic image
rendering from appearing too perfect, seemingly mechanical and
artif\/icial. For instance, it helps to give a vibrant appearance to
brush stroke rendering. Typically, when sampling strategies yield
centroidal regions, they also tend to produce homogeneous
conf\/igurations and vice versa, illustrating an apparent trade-of\/f
between these competing objectives. Heterogeneous conf\/igurations
are assessed using the Voronoi diagrams (Fig.~\ref{Fig8}) and mosaic
renderings (Fig.~\ref{Fig11}). Their ef\/fect is also visible in the
arrangement of brush strokes in non-photorealistic image rende\-ring
(Fig.~\ref{Fig12}). Random sampling produces the most heterogeneous local
conf\/igurations, as its mosaic tiles exhibit the greatest variety
of shapes. Jittered and quasirandom sampling are nearly as
heterogeneous, since their mosaic tiles are almost as widely
varied, though few of them are exceptionally large in size.
Quasicrystal and farthest point sampling are far less
heterogeneous. By upholding local symmetries, quasicrystal
sampling causes sample sites to have only a few possible local
conf\/igurations, resulting in mosaic tiles that have only a few
possible shapes. By upholding nearest neighbor distance, farthest
point sampling causes sample sites to have similar local
conf\/igurations, resulting in mosaic tiles that look very much
alike, mostly convex and rounded. Based on repetitions of a single
local conf\/iguration, periodic sampling is entirely homogeneous.
\end{enumerate}

\begin{figure}[t]
\centering
\setlength{\tabcolsep}{3pt}
\renewcommand{\arraystretch}{1.2}
{\footnotesize
\begin{tabular} {l|c|c|c|c|c| c|c }
Sampling  &  Accurate  &  Progressive &  Uniform  & Isotropic &
Blue Noise  & Centroidal  &  Heteroeneous
\\
Strategies & Reconstruction & Ref\/inement & Coverage & Distribution
& Spectrum & Regions & Conf\/igurations \\\hline\hline
 \rule{0pt}{12pt}Periodic &   ${\bigstar}\,{\bigstar}\,{\bigstar}\,{\bigstar}$ & { ${\bigstar}$} &  { ${\bigstar}\,{\bigstar}\,{\bigstar}\,{\bigstar}$} & { ${\bigstar}$} & { ${\bigstar}$} & { ${\bigstar}\,{\bigstar}\,{\bigstar}\,{\bigstar}$} & { ${\bigstar}$}\\
 Quasicrystal &  { ${\bigstar}\,{\bigstar}\,{\bigstar}$} & { ${\bigstar}\,{\bigstar}\,{\bigstar}\,{\bigstar}$} &  { ${\bigstar}\,{\bigstar}\,{\bigstar}$} & { ${\bigstar}$} & { ${\bigstar}$} & { ${\bigstar}\,{\bigstar}\,{\bigstar}$} & { ${\bigstar}\,{\bigstar}$}\\
 Farthest Point &  { ${\bigstar}\,{\bigstar}\,{\bigstar}\,{\bigstar}$} & { ${\bigstar}\,{\bigstar}\,{\bigstar}\,{\bigstar}$} &  { ${\bigstar}\,{\bigstar}\,{\bigstar}\,{\bigstar}$} & { ${\bigstar}\,{\bigstar}\,{\bigstar}$} & { ${\bigstar}\,{\bigstar}\,{\bigstar}\,{\bigstar}$} & { ${\bigstar}\,{\bigstar}\,{\bigstar}$} & { ${\bigstar}\,{\bigstar}$}\\
 Jittered &  { ${\bigstar}$} & { ${\bigstar}$} &  { ${\bigstar}\,{\bigstar}$} & { ${\bigstar}\,{\bigstar}\,{\bigstar}$} & { ${\bigstar}\,{\bigstar}\,{\bigstar}$} & { ${\bigstar}\,{\bigstar}$} & { ${\bigstar}\,{\bigstar}\,{\bigstar}$}\\
 Quasirandom&  { ${\bigstar}\,{\bigstar}$} & { ${\bigstar}\,{\bigstar}\,{\bigstar}$} &  { ${\bigstar}\,{\bigstar}$} & { ${\bigstar}\,{\bigstar}$} & { ${\bigstar}\,{\bigstar}$} & { ${\bigstar}\,{\bigstar}$} & { ${\bigstar}\,{\bigstar}\,{\bigstar}$}\\
 Random &  { ${\bigstar}$} & { ${\bigstar}\,{\bigstar}$} &  { ${\bigstar}$} & { ${\bigstar}\,{\bigstar}\,{\bigstar}\,{\bigstar}$} & {\centering ${\bigstar}$} & { ${\bigstar}$} & { ${\bigstar}\,{\bigstar}\,{\bigstar}\,{\bigstar}$}\\
\end{tabular}

\bigskip
${\bigstar}\,{\bigstar}\,{\bigstar}\,{\bigstar}$ Superior\quad
${\bigstar}\,{\bigstar}\,{\bigstar}$ Good\quad
${\bigstar}\,{\bigstar}$ Fair\quad ${\bigstar}$ Poor }
\caption{Qualitative evaluation of image sampling strategies.}\label{Fig10}
\end{figure}

Our qualitative analysis (Fig.~\ref{Fig10}) indicates that quasicrystal
sampling of\/fers a useful compromise between the ordered behavior
of standard periodic sampling using a regular square lattice and
the disordered behavior of standard Monte Carlo sampling using
jittered, quasirandom, or random sampling. Compared with periodic
sampling, quasicrystal sampling displays a~greater variety of
local sample site conf\/igurations resulting in smoother progressive
ref\/inement, although its sampling patterns are somewhat less
uniform, leading to lower accuracy of image reconstruction.
Compared with jittered, quasirandom, and random sampling,
quasicrystal sampling displays more uniform coverage resulting in
better accuracy of image reconstruction, although its sampling
patterns are anisotropic, exhibiting signif\/icantly less variety of
local sample site conf\/igurations. By virtue of its deterministic
construction, quasicrystal sampling does not suf\/fer from the
variability that can af\/fect the results of random sampling,
jittered sampling, and, to a much lesser degree, farthest point
sampling. Nevertheless, its lack of a blue noise power spectrum
renders it rather susceptible to aliasing artifacts. Research on
quasicrystal sampling based on the Penrose tiling~\cite{Fast
hierarchical importance sampling with blue noise properties}
suggests that it may be possible to partially alleviate this
problem by taking advantage of the symmetries and the repetitions
of the local sample site conf\/igurations in order to systematically
displace the sample sites in a manner that improves the spectral
properties of the sampling pattern.

\looseness=2
Based on our qualitative evaluation of the various non-adaptive
sampling strategies (Fig.~\ref{Fig10}), we recommend a blue noise
sampling strategy, such as farthest point sampling, for general
use in image representation. In particular, farthest point
sampling does not perform poorly on any of our seven evaluation
criteria. Overall, our qualitative evaluation of non-adaptive
sampling strategies is in broad agreement with previous studies,
which did not consider quasicrystal sampling. They emphasized the
importance of Poisson disk distributions~\cite{Principles of
digital image synthesis} and low discrepancy distributions
\cite{Discrepancy as a quality measure for sample distributions},
which are exemplif\/ied in our evaluation by farthest point sampling
and quasirandom sampling respectively. Hence, the good overall
performance of these two techniques should come as no surprise.
Farthest point sampling performs better than quasirandom sampling
on six out of the seven evaluation criteria. For the majority of
our evaluation criteria, quasicrystal sampling performs no better
than farthest point sampling and no worse than quasirandom
sampling. Nevertheless, from a practical point of view,
quasicrystal sampling is signif\/icantly simpler to implement and
calculate than farthest point sampling, which relies on
maintaining complex geometric data structures to keep track of the
vertices of a Voronoi diagram. This could be an important
consideration for imaging applications on mobile devices that have
limited processing and storage capabilities. From a theoretical
point of view, the deterministic algebraic construction of
quasicrystals renders their sampling patterns particularly well
suited to mathematical analysis. Presenting possibilities for future research, the cut-and-project method could be adapted for higher dimensional
sampling or adaptive sampling applications.

\looseness=1
In future work, it would also be interesting to explore the
relationship between local symmetry and sampling quality. The
cut-and-project method can be used to generate non-periodic point
sets with dif\/ferent symmetries, not just the pentagonal and
decagonal symmetries associated with the golden ratio, as shown in
this work. Just as for periodic sampling it would be interesting
to compare the image reconstruction accuracy of square and
hexagonal grids, for non-periodic sampling it would be interesting
to compare our decagonal quasicrystal tiling with the dodecagonal
Socolar tiling, which was recently proposed for use in sampling
applications~\cite{BuildingLowDiscrepancy}.

\section{Conclusion}

\looseness=1
Cut-and-project quasicrystals present new possibilities for image
sampling in computer gra\-phics. This non-periodic sampling approach
deterministically generates uniformly space-f\/illing point sets,
ensuring that sample sites are evenly distributed throughout the
image. It of\/fers a useful compromise between predictability and
randomness, between the standard periodic sampling and the
standard Monte Carlo sampling methods. Although blue noise
sampling can generate higher quality sampling patterns for
photorealistic image reconstruction, quasicrystal sampling may
prove much simpler to implement and calculate. In the context of
non-photorealistic image rendering, quasicrystal sampling may
prove attractive for its symmetry properties.

\begin{figure}[p]
\setlength{\tabcolsep}{7pt}
\centering
\small
\begin{tabular}{cc}
\includegraphics[width=70mm]{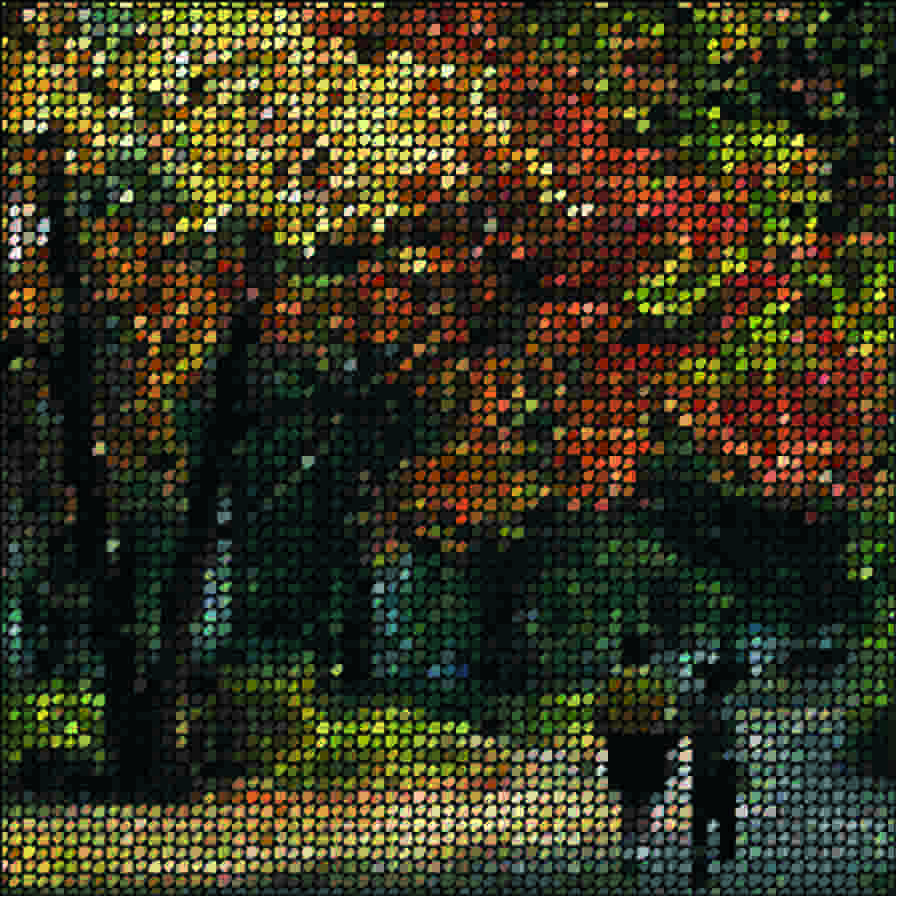}
&
\includegraphics[width=70mm]{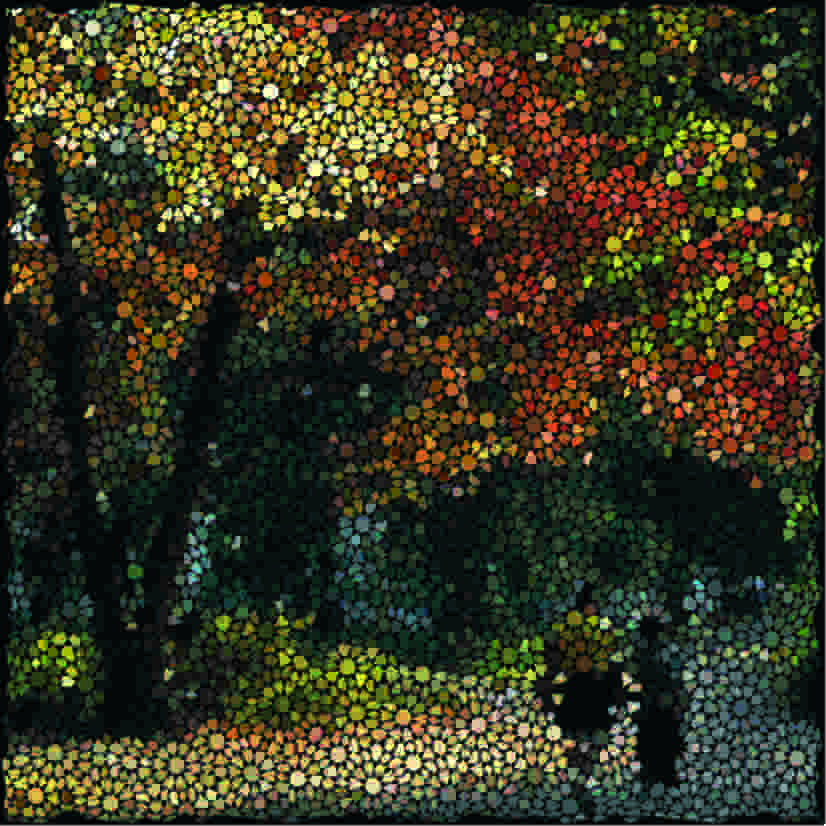}
\\
Periodic sampling & Quasicrystal sampling \\[2mm]
\includegraphics[width=70mm]{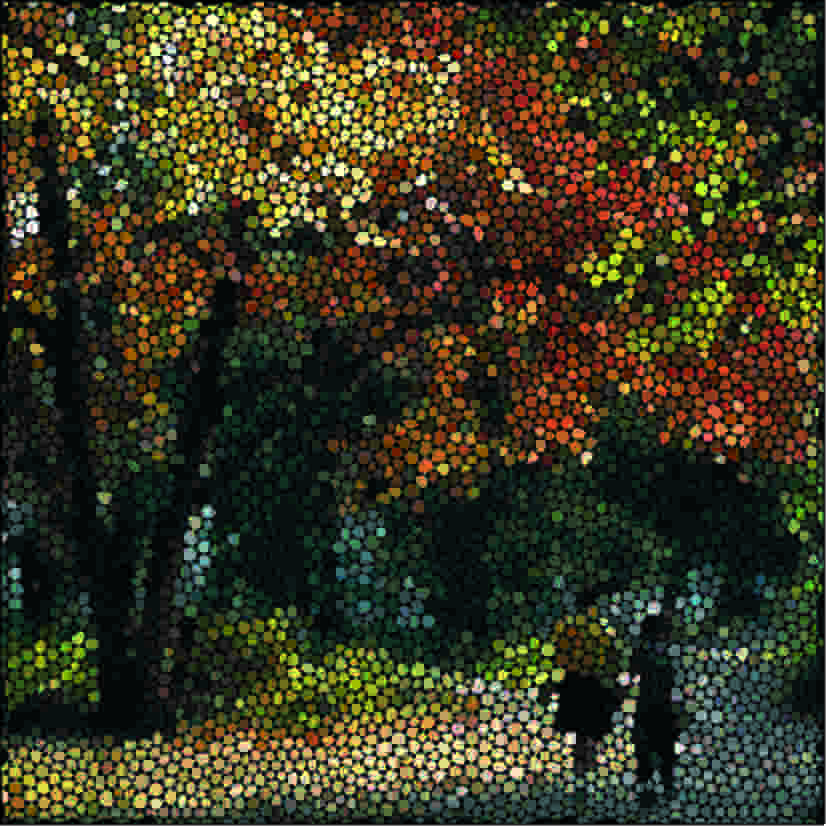}
&
\includegraphics[width=70mm]{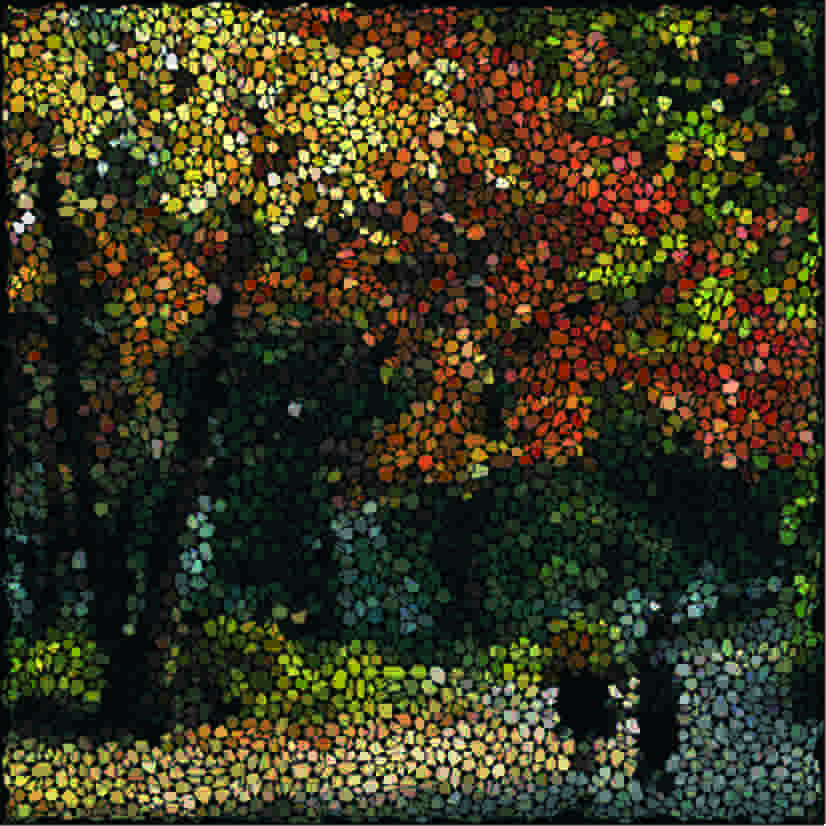}
\\
Farthest point sampling & Jittered sampling \\[2mm]
\includegraphics[width=70mm]{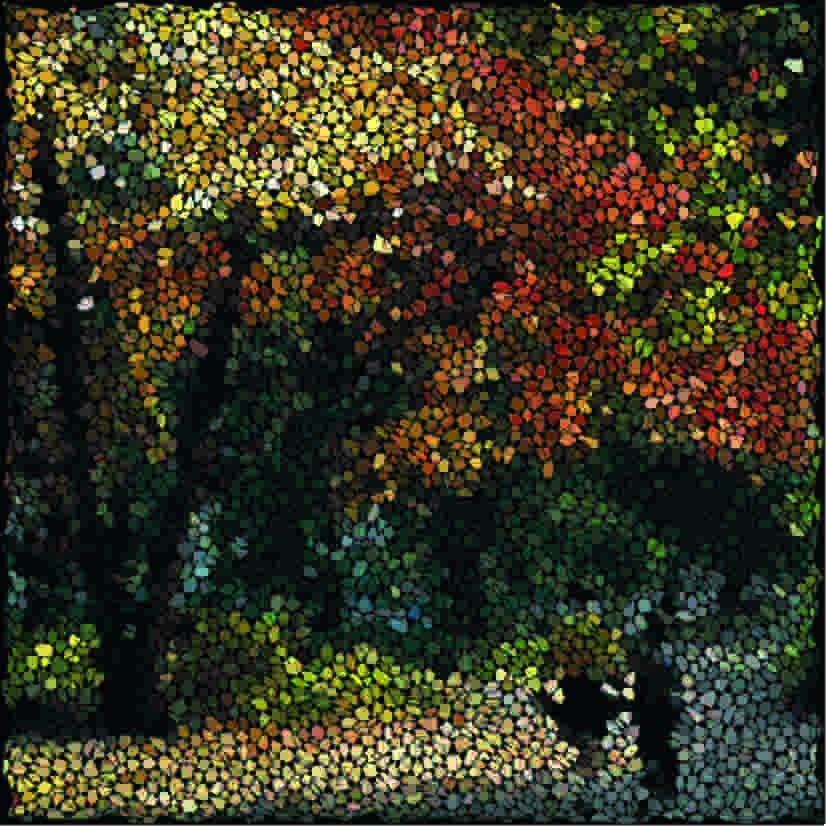}
&
\includegraphics[width=70mm]{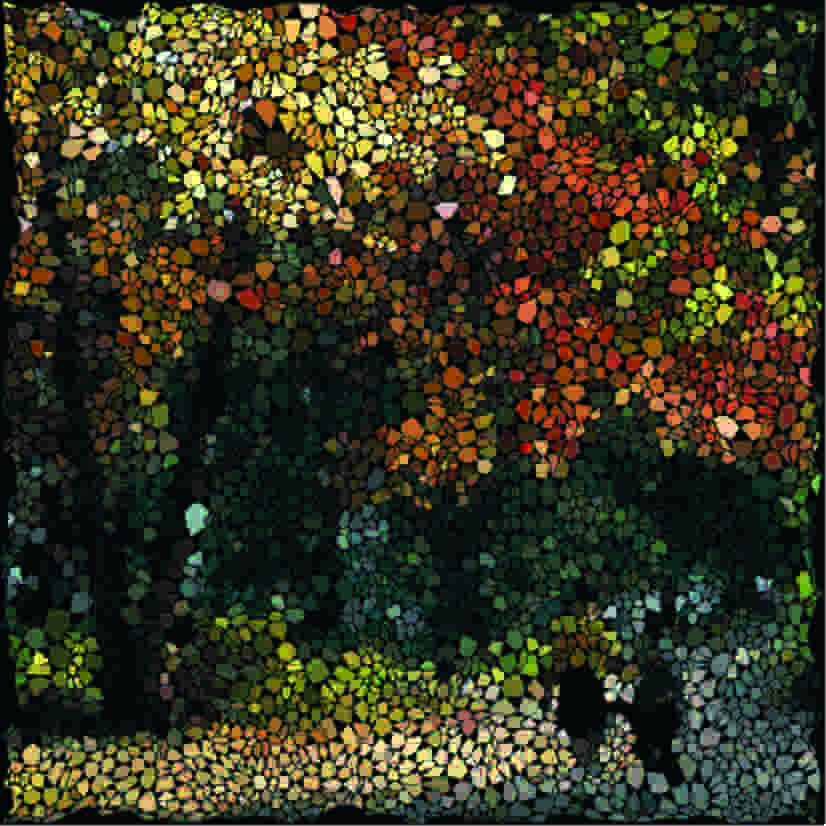}
\\
Quasirandom sampling & Random sampling
\end{tabular}
\caption{Image sampling strategies rendered using the mosaic
style (4225 samples $\approx 2.6\%$).}\label{Fig11}
\end{figure}

\begin{figure}[p]
\setlength{\tabcolsep}{7pt}
\centering
\small
\begin{tabular}{cc}
\includegraphics[width=70mm]{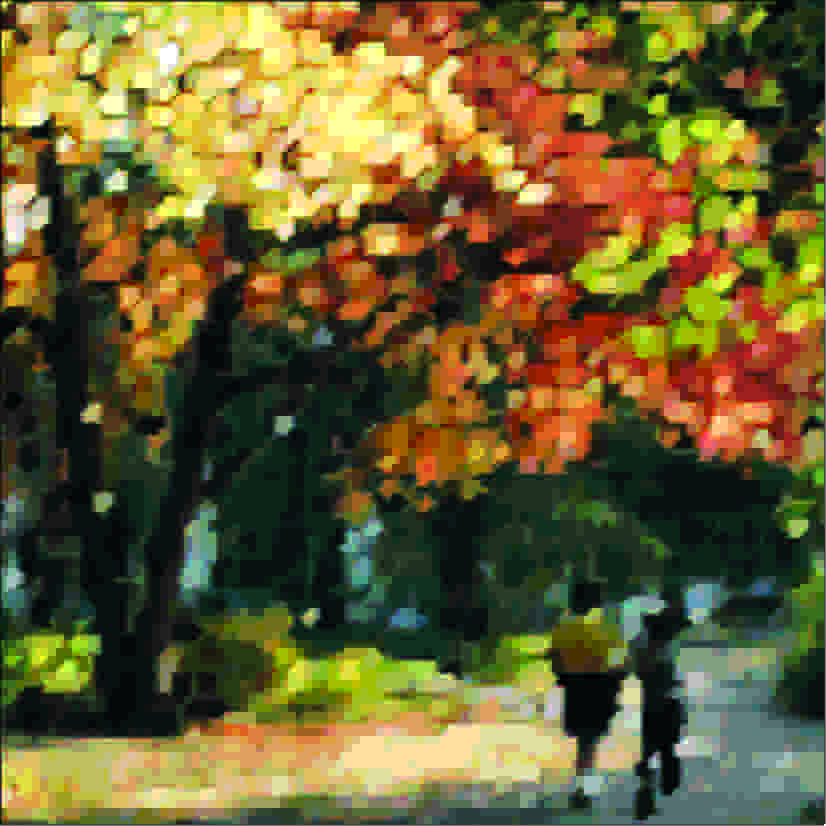}
&
\includegraphics[width=70mm]{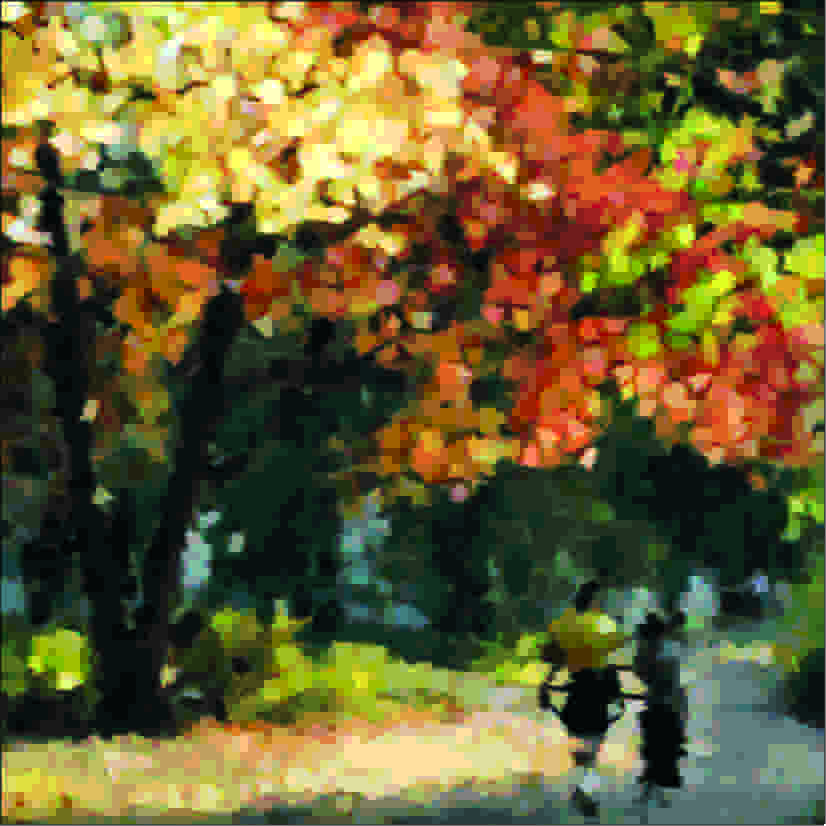}
\\
Periodic sampling & Quasicrystal sampling \\[2mm]
\includegraphics[width=70mm]{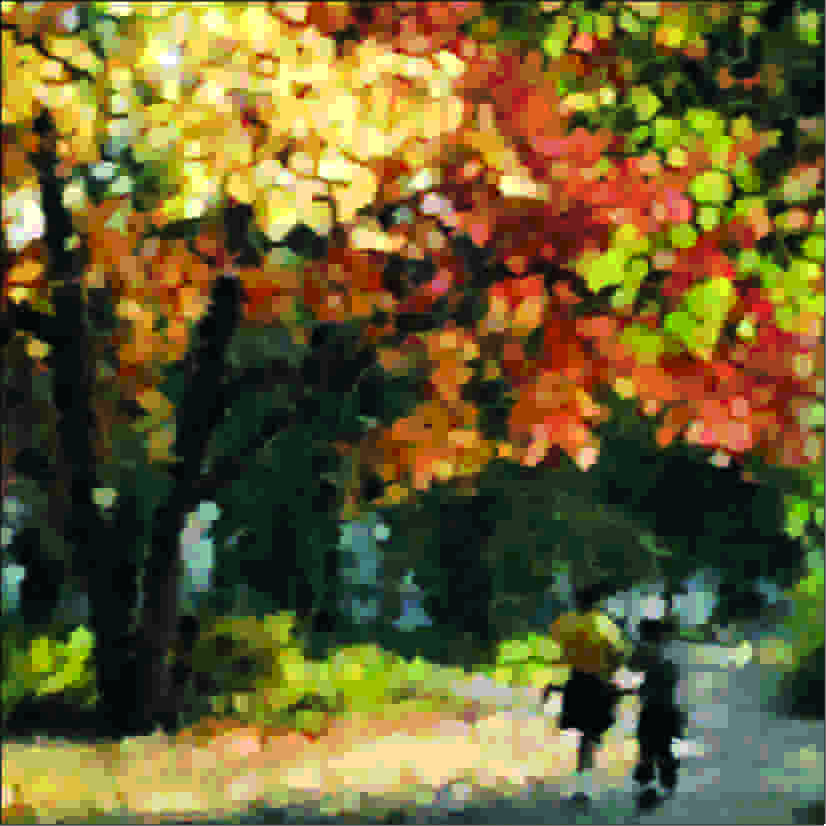}
&
\includegraphics[width=70mm]{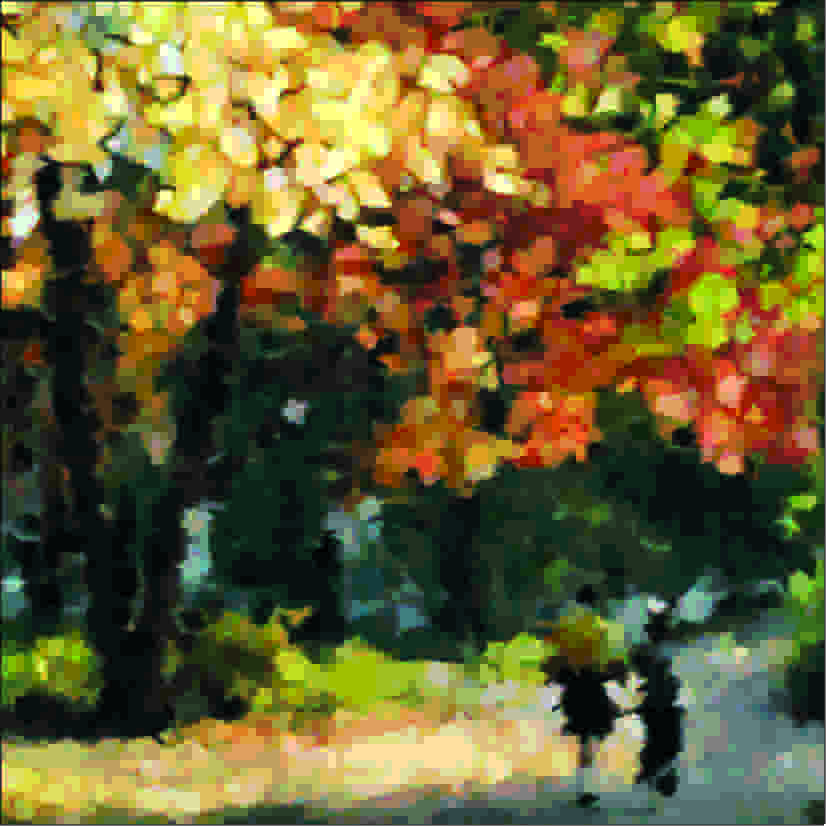}
\\
Farthest point sampling & Jittered sampling \\[2mm]
\includegraphics[width=70mm]{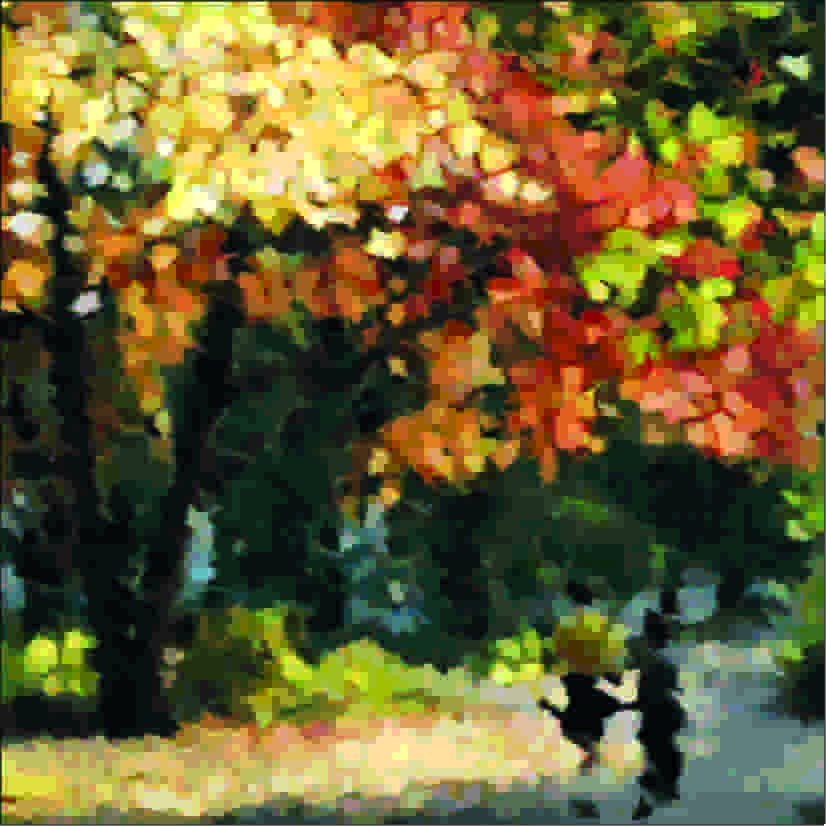}
&
\includegraphics[width=70mm]{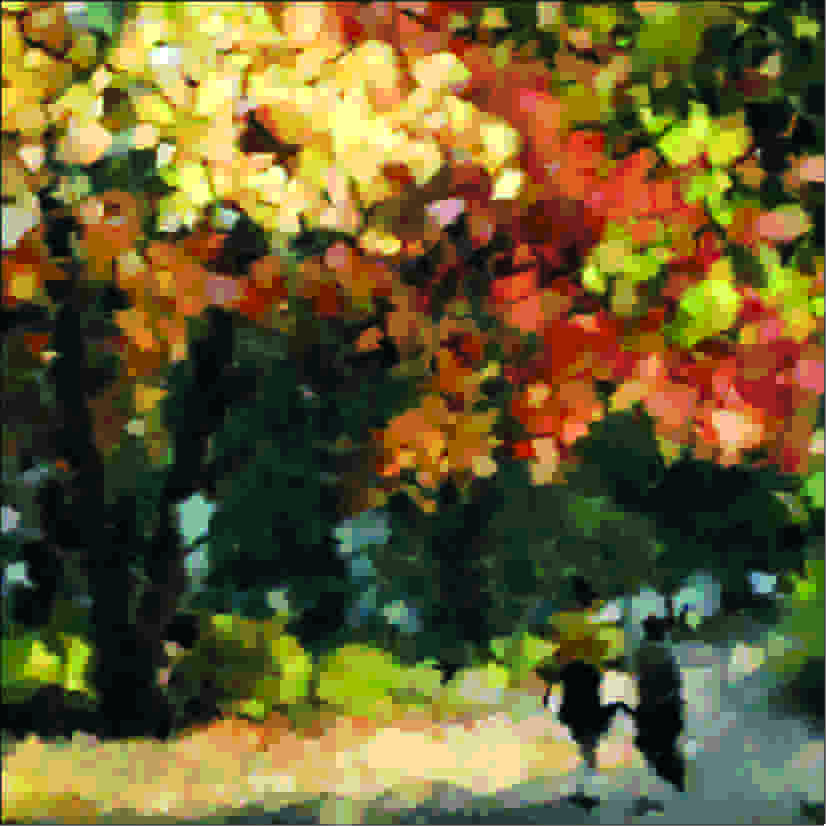}
\\
Quasirandom sampling & Random sampling
\end{tabular}
\caption{Image sampling strategies rendered using the ``paint
strokes'' style (4225 samples $\approx 2.6\%$).}\label{Fig12}
\end{figure}

\begin{figure}[p]
\setlength{\tabcolsep}{7pt}
\centering
\small
\begin{tabular}{cc}
\includegraphics[width=70mm]{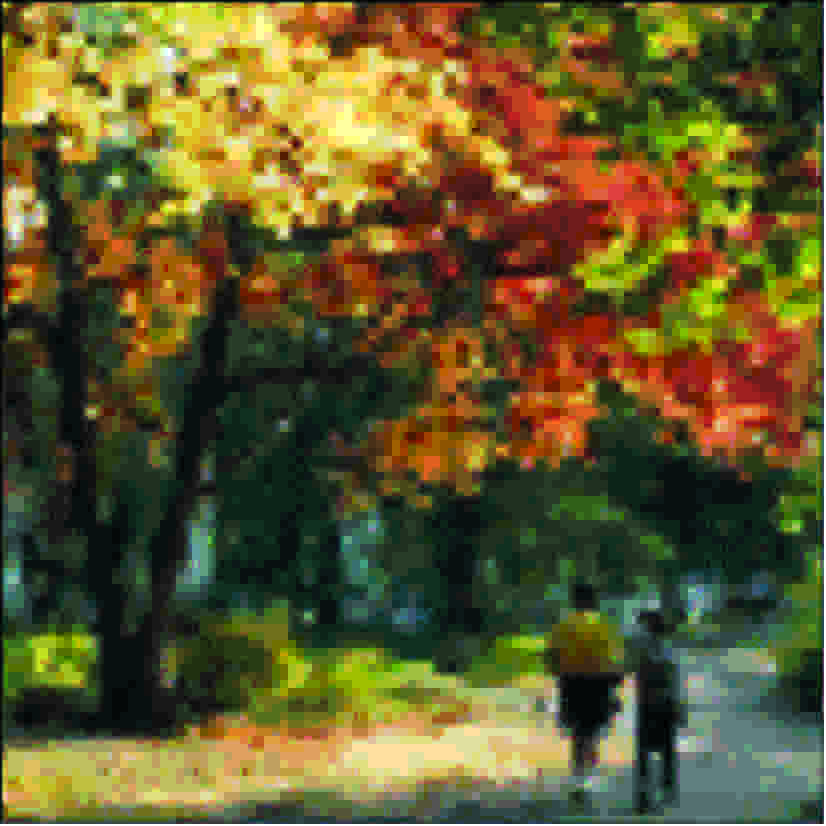}
&
\includegraphics[width=70mm]{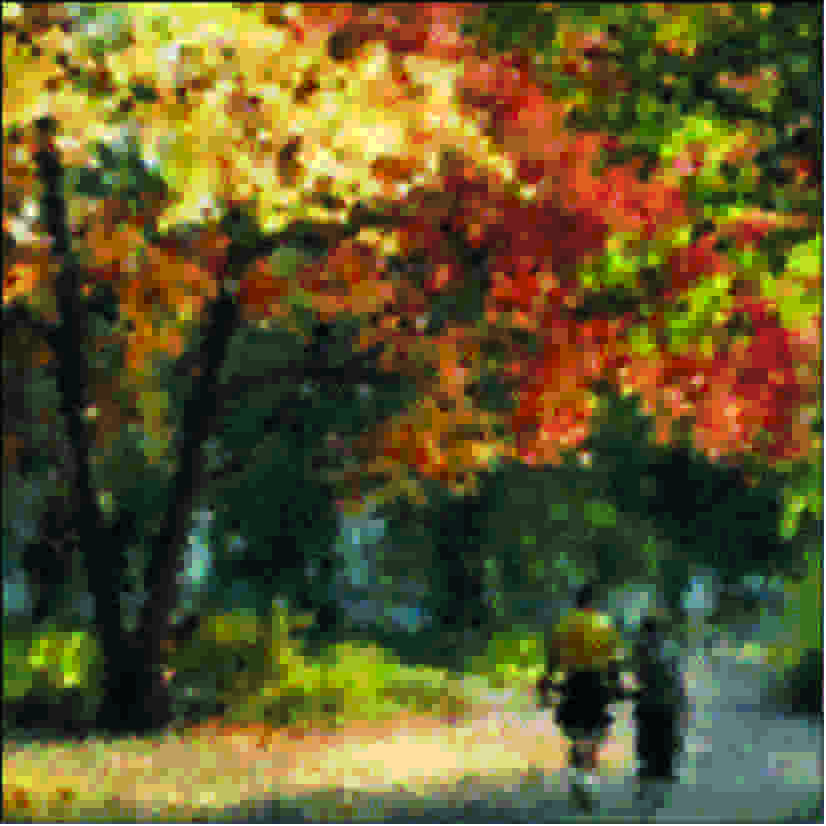}
\\
Periodic sampling (PSNR $= 18.58$)
& Quasicrystal sampling (PSNR $= 18.52$) \\[2mm]
\includegraphics[width=70mm]{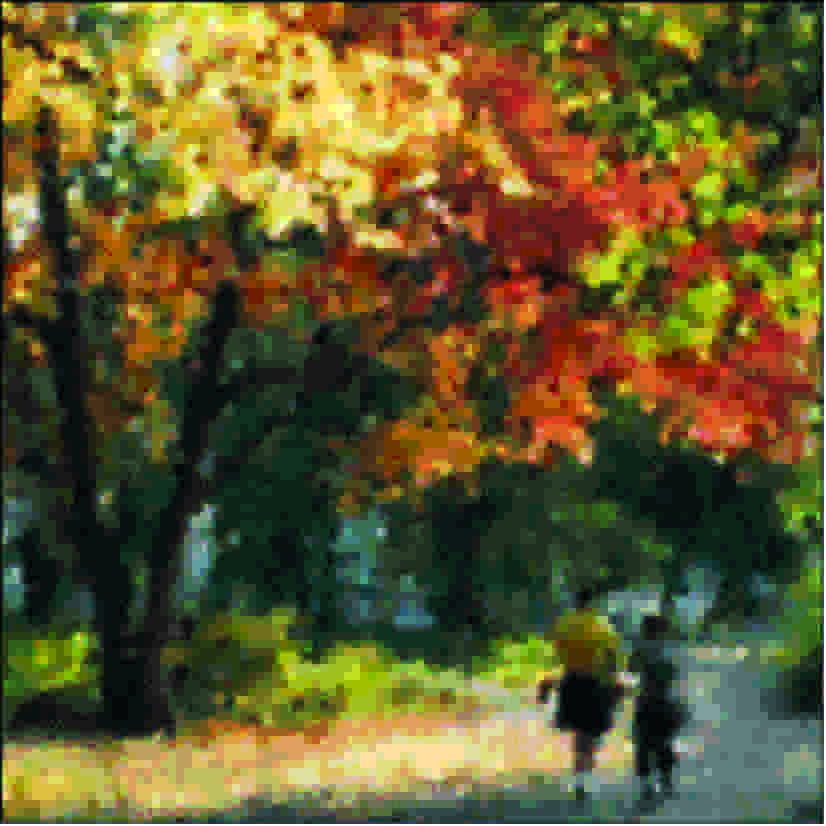}
&
\includegraphics[width=70mm]{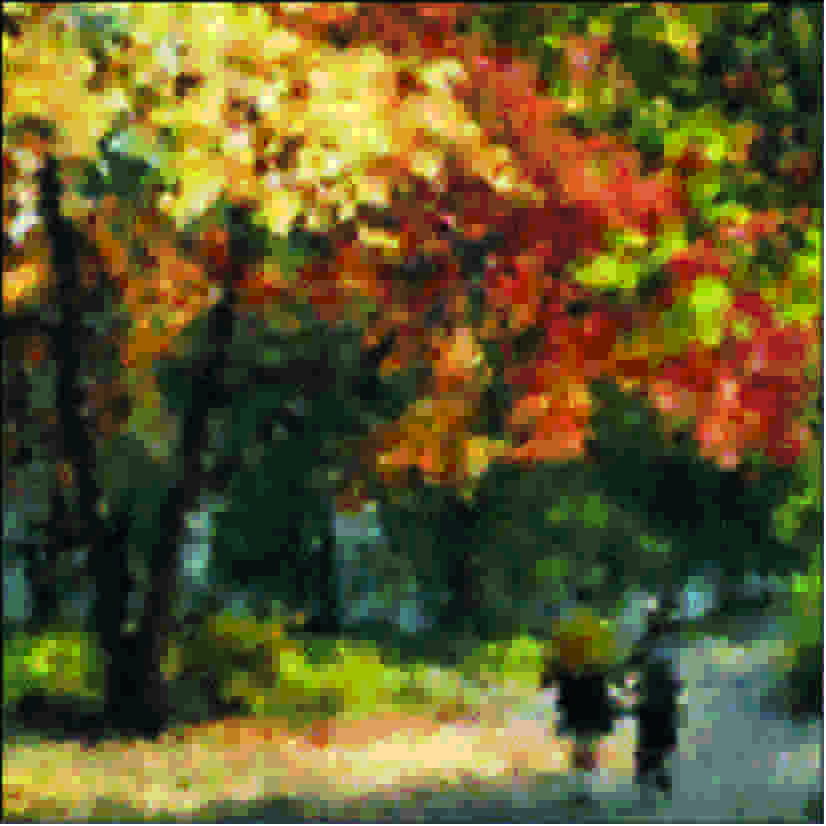}
\\
Farthest point sampling (PSNR $= 18.55$)
& Jittered sampling (PSNR $= 18.31$) \\[2mm]
\includegraphics[width=70mm]{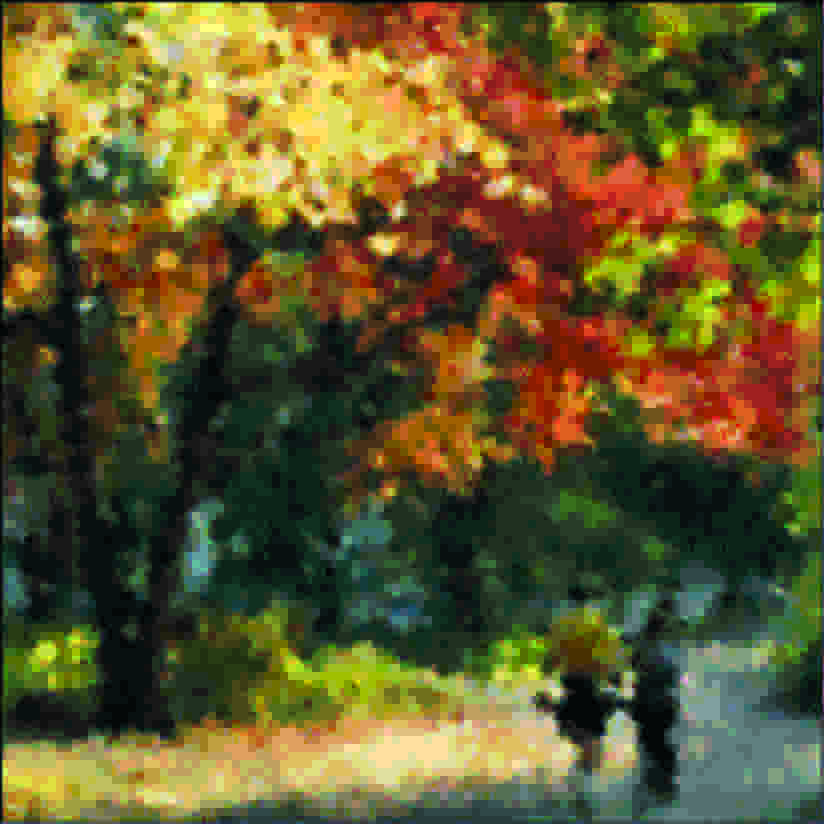}
&
\includegraphics[width=70mm]{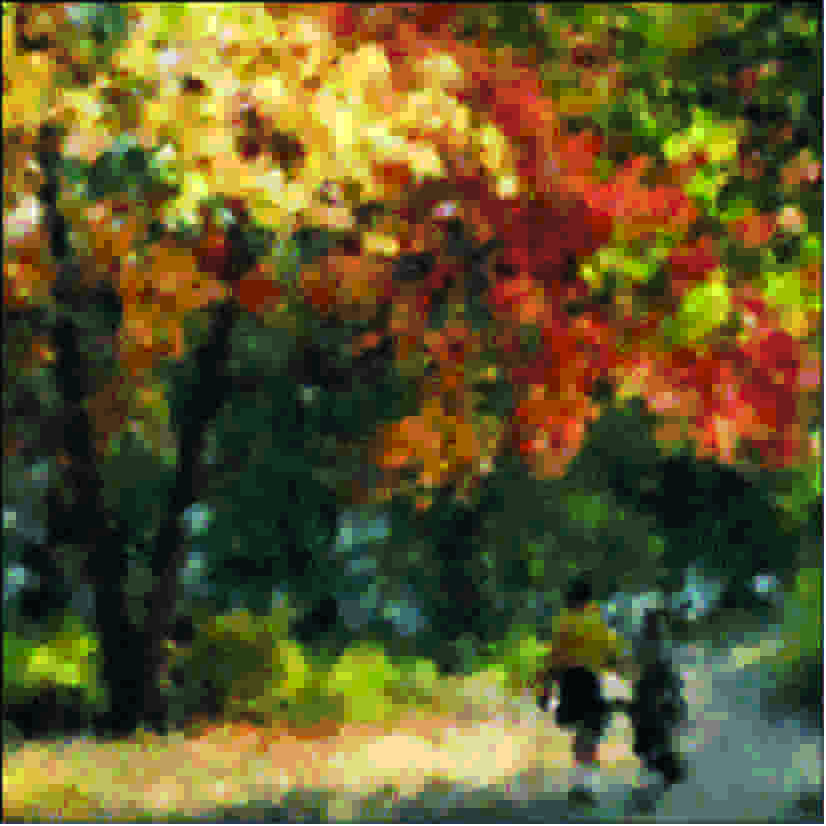}
\\
Quasirandom sampling (PSNR $= 18.24$)
& Random sampling (PSNR $= 17.93$)
\end{tabular}
 \caption{Image sampling strategies rendered using Shepard
interpolation (4225 samples $\approx 2.6\%$).}\label{Fig13}
\end{figure}

\begin{figure}[p]
\centering
\includegraphics[scale=0.95]{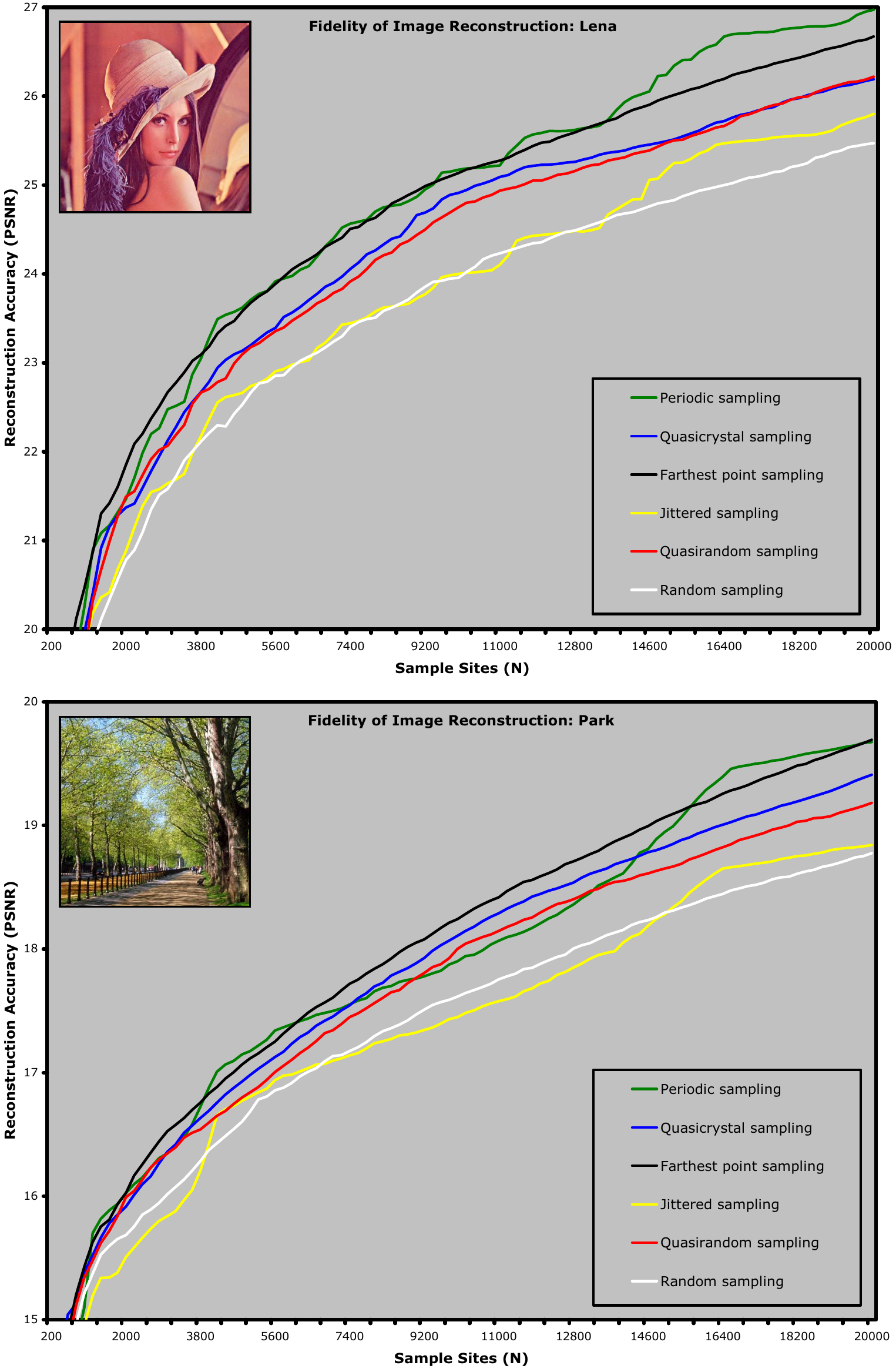}
\caption{Quantitative evaluation of image sampling strategies rendered using Gouraud shading.}\label{Fig14}
\end{figure}

\subsection*{Acknowledgements}

\looseness=1
We are grateful to the referees for valuable remarks that led to improvements of the original
manuscript. We also acknowledge the f\/inancial support of the Natural Sciences and
Engineering Research Council of Canada, le Fonds Qu\'eb\'ecois de la
Recherche sur la Nature et les Technologies, as well as the grants
MSM6840770039 and LC06002 of the Ministry of Education of the
Czech Republic, and GA201/09/0584 of the Czech Science Foundation.
We are also grateful for the support of the MIND Research
Institute and the Merck Frosst Company. Mark Grundland further
acknowledges the f\/inancial assistance of the Celanese Canada
Internationalist Fellowship, the British Council, the Cambridge
Commonwealth Trust, and the Overseas Research Student Award
Scheme. The images were provided by FreeFoto.com and the Waterloo
Brag Zone.
\newpage

\section*{Supplementary materials}

 For a full resolution version of this paper, along with supplementary materials, please visit:
\url{http://www.Eyemaginary.com/Portfolio/Publications.html}.

\pdfbookmark[1]{References}{ref}
\LastPageEnding


\begin{thebibliography}{99}

\footnotesize\itemsep=0pt

\bibitem{ScatteredData}
Amidror I.,
Scattered data interpolation methods for electronic imaging systems: a survey,
\emph{J. Elec\-tronic Imaging} {\bf 11} (2002), 157--176.


\bibitem{Non-crystallographic root systems}
Chen L., Moody R.V., Patera J.,
Non-crystallographic root systems,
in  Quasicrystals and Discrete Geometry  (Toronto, ON, 1995), {\it Fields Inst. Monogr.}, Vol.~10, Amer. Math. Soc., Providence, RI, 1998, 135--178.


\bibitem{WangTiles}
Cohen M.F., Shade J., Hiller S., Deussen O.,
Wang tiles for image and texture generation,
in Proceedings of SIGGRAPH, 2003, 287--294.


\bibitem{StochasticSampling}
Cook R.L.,
Stochastic sampling in computer graphics,
{\it ACM Trans. Graphics} {\bf 5} (1986), 51--72.


\bibitem{FloatingPoints}
Deussen O., Hiller S., van Overveld C., Strothotte T.,
Floating points: a method for computing stipple drawings,
in Proceedings of EUROGRAPHICS, 2000, 41--50.


\bibitem{Antialiasing through stochastic sampling}
Dippe M.A.Z., Wold E.H.,
Antialiasing through stochastic sampling,
in Proceedings of SIGGRAPH, 1985, 69--78.


\bibitem{CentroidalVoronoiTesselations}
Du Q., Faber V., Gunzburger M.,
Centroidal Voronoi tessellations: applications and algorithms,
{\it SIAM Rev.} {\bf 41} (1999), 637--676.


\bibitem{SpatialDataStructuren}
Dunbar D., Humphreys G.,
A spatial data structure for fast Poisson-disk sample generation,
in  Proceedings of SIGGRAPH, 2006, 503--508.


\bibitem{FarthestPointStrategy}
Eldar Y., Lindenbaum M., Porat M., Zeevi Y.Y.,
The farthest point strategy for progressive image sampling,
{\it IEEE Trans. Image Process.} {\bf 6} (1997), 1305--1315.


\bibitem{Principles of digital image synthesis}
Glassner A.,
Principles of digital image synthesis, Vol.~1, Morgan Kaufmann, 1995.


\bibitem{Aperiodic tiling}
Glassner A.,
Aperiodic tiling,
{\it IEEE Computer Graphics and Applications} {\bf 18} (1998), no.~3, 83--90.


\bibitem{Penrose tiling}
Glassner A.,
Penrose tiling,
{\it IEEE Computer Graphics and Applications} {\bf 18} (1998), no.~4, 78--86.


\bibitem{Aperiodic hierarchical tilings}
Goodman-Strauss C.,
Aperiodic hierarchical tilings,
in  Foams, Emulsions, and Cellular Materials (Carg\`ese, 1997), {\it NATO Adv. Sci. Inst. Ser. E Appl. Sci.}, Vol.~354, Kluwer Acad. Publ., Dordrecht, 1999, 481--496.


\bibitem{Tilings and patterns}
Grunbaum B., Shephard G.C.,
Tilings and patterns, WH Freeman, 1987.


\bibitem{StyleAndContent}
Grundland M.,
 Style and content in digital imaging: Reconciling aesthetics with ef\/f\/iciency in image repre\-sentation, VDM, 2008.


\bibitem{Stylized multiresolution image representation}
Grundland M., Gibbs C., Dodgson N.A.,
Stylized multiresolution image representation,
{\it J. Electro\-nic Imaging} {\bf 17} (2008), 013009, 1--17.


\bibitem{Simulating decorative mosaics}
Hausner A.,
Simulating decorative mosaics,
in Proceedings of SIGGRAPH, 2001, 573--580.


\bibitem{Pointillist halftoning}
Hausner A.,
Pointillist halftoning,
in Proceedings of the International Conference on Computer Graphics and Imaging, 2005, 134--139.


\bibitem{Tiled blue noise samples}
Hiller S., Deussen O., Keller A.,
Tiled blue noise samples,
in Proceedings of Vision, Modeling and Visuali\-zation, 2001, 265--271.


\bibitem{BeyondStippling}
Hiller S., Hellwig H., Deussen O.,
Beyond stippling -- methods for distributing objects on the plane,
in  Proceedings of EUROGRAPHICS, 2003, 515--522.


\bibitem{EfficientGeneration}
Jones T.R.,
Ef\/f\/icient generation of Poisson-disk sampling patterns,
{\it J. Graphics Tools} {\bf  11} (2006), no.~2, 27--36.


\bibitem{Filtered jitter}
Klassen R.V.,
Filtered jitter,
{\it Computer Graphics Forum} {\bf 19} (2000), no.~4, 223--230.


\bibitem{Recursive wang tiles for real-time blue noise}
Kopf J., Cohen-Or D., Deussen O., Lischinski D.,
Recursive Wang tiles for real-time blue noise,
in  Pro\-ceedings of SIGGRAPH, 2006, 509--518.


\bibitem{A procedural object distribution function}
Lagae A., Dutre P.,
A procedural object distribution function,
{\it ACM Trans. Graphics} {\bf 24} (2005), 1442--1461.


\bibitem{AlternativeForWang}
Lagae A., Dutre P.,
An alternative for Wang tiles: colored edges versus colored corners,
{\it ACM Trans. Graphics} {\bf 25} (2006), 1442--1459.


\bibitem{ComparisonOfMethods}
Lagae A., Dutre P.,
A comparison of methods for generating Poisson disk distributions,
{\it Computer Graphics Forum} {\bf  27} (2008), no.~1, 114--129.


\bibitem{Tile-based methods for interactive applications}
Lagae A., Kaplan C.S., Fu C.-W., Ostromoukhov V., Deussen O.,
Tile-based methods for interactive applications,
SIGGRAPH 2008 Class Notes, ACM, 2008.

\bibitem{MedievalArchitecture}
Lu P.J., Steinhardt P.J.,
Decagonal and quasi-crystalline tilings in medieval Islamic architecture,
{\it Science} {\bf 315} (2007), no.~5815, 1106--1110.


\bibitem{ZichIII}
Mas\'akov\'a Z., Patera J., Zich J.,
Classif\/ication of Voronoi and Delone tiles of quasicrystals. III.~Decagonal acceptance window of any size,
{\it J. Phys. A: Math. Gen.} {\bf 38} (2005), 1947--1960.


\bibitem{Hierarchical poisson disk sampling distributions}
McCool M., Fiume E.,
Hierarchical Poisson disk sampling distributions,
in  Proceedings of Graphics Interface, 1992, 94--105.


\bibitem{Algebraic numbers and harmonic analysis}
Meyer Y.,
Algebraic numbers and harmonic analysis, North-Holland, 1972.


\bibitem{Spectrally optimal sampling for distribution ray tracing}
Mitchell D.P.,
Spectrally optimal sampling for distribution ray tracing,
in  Proceedings of SIGGRAPH, 1991, 157--164.


\bibitem{Color quantization and processing by fibonacci lattices}
Mojsilovic A., Soljanin E.,
Color quantization and processing by Fibonacci lattices,
{\it IEEE Trans. Image Process.} {\bf 10} (2001), 1712--1725.


\bibitem{Quasicrystals and icosians}
Moody R.V., Patera J.,
Quasicrystals and icosians,
{\it J. Phys. A: Math. Gen.} {\bf 26} (1993), 2829--2853.


\bibitem{Dynamical generation of quasicrystals}
Moody R.V., Patera J.,
Dynamical generation of quasicrystals,
{\it Lett. Math. Phys.} {\bf 36} (1996), 291--300.


\bibitem{Mathematical tools for computer-generated ornamental patterns}
Ostromoukhov V.,
Mathematical tools for computer-generated ornamental patterns,
{\it Lecture Notes in Computer Science}, Vol.~1375, 1998, 193--223.

\bibitem{Fast hierarchical importance sampling with blue noise properties}
Ostromoukhov V., Donohue C., Jodoin P.M.,
Fast hierarchical importance sampling with blue noise pro\-per\-ties,
in Proceedings of SIGGRAPH, 2004, 488--495.

\bibitem{BuildingLowDiscrepancy}
Ostromoukhov V.,
 Building 2D low-discrepancy sequences for hierarchical importance sampling using dode\-cagonal aperiodic tiling,
in Proceedings of GRAPHICON, 2007, 139--142.

\bibitem{Sampling with polyominoes}
Ostromoukhov V.,
Sampling with polyominoes,
in  Proceedings of SIGGRAPH, 2007, 078, 1--6.

\bibitem{Non-crystallographic root systems and quasicrystals}
Patera J.,
Non-crystallographic root systems and quasicrystals.
in  The Mathematics of Long-Range Aperio\-dic Order  (Waterloo, ON, 1995), {\it NATO Adv. Sci. Inst. Ser. C Math. Phys. Sci.}, Vol.~489, Kluwer Acad. Publ., Dordrecht, 1997, 443--465.


\bibitem{Numerical recipes in c}
Press W., Teukolsky S.A., Vetterling W.T., Flannery B.P.,
Numerical recipes in C, 2nd ed., Cambridge University Press, 1992.


\bibitem{Computer generation of penrose tilings}
Rangel-Mondragon J., Abas S.J.,
Computer generation of Penrose tilings,
{\it Computer Graphics Forum} {\bf 7} (1988), no.~1, 29--37.


\bibitem{Weighted voronoi stippling}
Secord A.,
Weighted Voronoi stippling,
in  Proceedings of the Second International Symposium on Non-Photorealistic Animation and Rendering, 2002, 37--43.


\bibitem{Quasicrystals and geometry}
Senechal M.,
Quasicrystals and geometry, Cambridge University Press, Cambridge, 1995.


\bibitem{Digital color imaging handbook}
Sharma G.,
Digital color imaging handbook, CRC Press, 2003.


\bibitem{Schechtman}
Shechtman D., Blech I., Gratias D., Cahn J.W.,
Metallic phase with long-range orientational order and no translational symmetry,
{\it Phys. Rev. Lett.} {\bf 53} (1984), 1951--1953.


\bibitem{Discrepancy as a quality measure for sample distributions}
Shirley P.,
Discrepancy as a quality measure for sample distributions,
in Proceedings of EUROGRAPHICS, 1991, 183--194.


\bibitem{Aperiodic texture mapping}
Stam J.,
Aperiodic texture mapping,
European Research Consortium for Informatics and Mathematics, Technical Report ERCIM-01/97-R046, 1997.


\bibitem{Tile-based texture mapping on graphics hardware}
Wei L.-Y.,
Tile-based texture mapping on graphics hardware,
in Proceedings of the ACM Conference on Graphics Hardware, 2004, 55--63.


\bibitem{Parallel poisson disk sampling}
Wei L.-Y.,
Parallel Poisson disk sampling,
in Proceedings of SIGGRAPH, 2008, 020, 1--10.


\bibitem{Poisson disk point sets by hierarchical dart throwing}
White K.B., Cline D., Egbert P.K.,
Poisson disk point sets by hierarchical dart throwing,
in Proceedings of the IEEE Symposium on Interactive Ray Tracing, 2007, 129--132.

\end{thebibliography}
\end{document}